%% file: camera_ready.tex
\documentclass[twoside]{article}
\usepackage[accepted]{aistats2022}

\usepackage{todonotes}

\usepackage[round]{natbib}

\usepackage[utf8]{inputenc}
\usepackage[T1]{fontenc}   
\usepackage{hyperref}      
\usepackage{url}           
\usepackage{amsfonts}      
\usepackage{nicefrac}      
\usepackage{microtype}     
\usepackage{xcolor}        
\usepackage{todonotes}
\usepackage{amsthm, amsfonts, amssymb, amsmath}
\usepackage{enumitem}
\usepackage{mathabx}
\usepackage{setspace}
\usepackage{mathtools}
\usepackage{mathdots}
\usepackage{graphicx}
\usepackage{subcaption}
\usepackage{mwe}
\usepackage{longtable}

\usepackage[tableposition=top, justification=centering]{caption}

\usepackage{booktabs}
\usepackage{multirow, makecell}

\usepackage{algorithm2e}
\usepackage{lipsum}

\theoremstyle{plain}
\newtheorem{theorem}{Theorem}
\newtheorem{corollary}[theorem]{Corollary}
\newtheorem{lemma}[theorem]{Lemma}

\theoremstyle{definition}
\newtheorem{definition}{Definition}
\newtheorem*{definition*}{Definition}

\makeatletter
\def\blfootnote{\gdef\@thefnmark{}\@footnotetext}
\makeatother

\SetKwInput{KwInput}{Input} 
\SetKwInput{KwOutput}{Output}

\input macros

\begin{document}

\runningauthor{TaeHo Yoon, Youngsuk Park, Ernest K.\ Ryu, Yuyang Wang}

\twocolumn[

\aistatstitle{Robust Probabilistic Time Series Forecasting}

\aistatsauthor{ 
TaeHo Yoon\textsuperscript{\dag}\textsuperscript{*}
\And Youngsuk Park\textsuperscript{\ddag}
\And Ernest K. Ryu\textsuperscript{\dag}
\And Yuyang Wang\textsuperscript{\ddag}
}

\aistatsaddress{
\textsuperscript{\dag}Department of Mathematical Sciences, Seoul National University \\
\textsuperscript{\ddag}AWS AI Labs, Amazon Research
} 

]

\begin{abstract}
Probabilistic time series forecasting has played critical role in decision-making processes due to its capability to quantify uncertainties.
Deep forecasting models, however, could be prone to input perturbations, and the notion of such perturbations, together with that of robustness, has not even been completely established in the regime of probabilistic forecasting.
In this work, we propose a framework for robust probabilistic time series forecasting.
First, we generalize the concept of adversarial input perturbations, based on which we formulate the concept of robustness in terms of bounded Wasserstein deviation.
Then we extend the randomized smoothing technique to attain robust probabilistic forecasters with theoretical robustness certificates against certain classes of adversarial perturbations.
Lastly, extensive experiments demonstrate that our methods are empirically effective in enhancing the forecast quality under additive adversarial attacks and forecast consistency under supplement of noisy observations.
The code for our experiments is available at \url{https://github.com/tetrzim/robust-probabilistic-forecasting}.
\end{abstract}

\section{INTRODUCTION}
\label{section:introduction}

Time series forecasting is among the most important tasks in the automation and optimization of business processes. 
In retail, for example, determining how many units of each item to purchase and where to store them depends on forecasts of future demand over different regions.
In cloud computing, the estimated future usage of services and infrastructure components guides capacity planning \citep{park2019linear, park2020structured}. 
The real-time forecasting is essential as a subroutine for vehicle control and planning \citep{kim2020optimal} and other numerous applications \citep{petropoulos2021forecasting}. 
Due to its crucial role in downstream decision making, there are two desirable properties of a forecaster: 1) the ability to generate probabilistic forecasts that allows for uncertainty estimation; 2) reliability, in the sense of being robust to (potentially adversarial) input perturbations. 
In the present work, we investigate robust probabilistic forecasting models which aim to satisfy the both requirements.

\blfootnote{\hspace{-.2cm}\textsuperscript{\textbf{*}}Work done as an intern at Amazon Research. Correspondence to: TaeHo Yoon <\href{mailto:tetrzim@snu.ac.kr}{tetrzim@snu.ac.kr}>, Youngsuk Park <\href{mailto:pyoungsu@amazon.com}{pyoungsu@amazon.com}>.}

\begin{figure}[t]
    \centering
    \includegraphics[width=.45\textwidth]{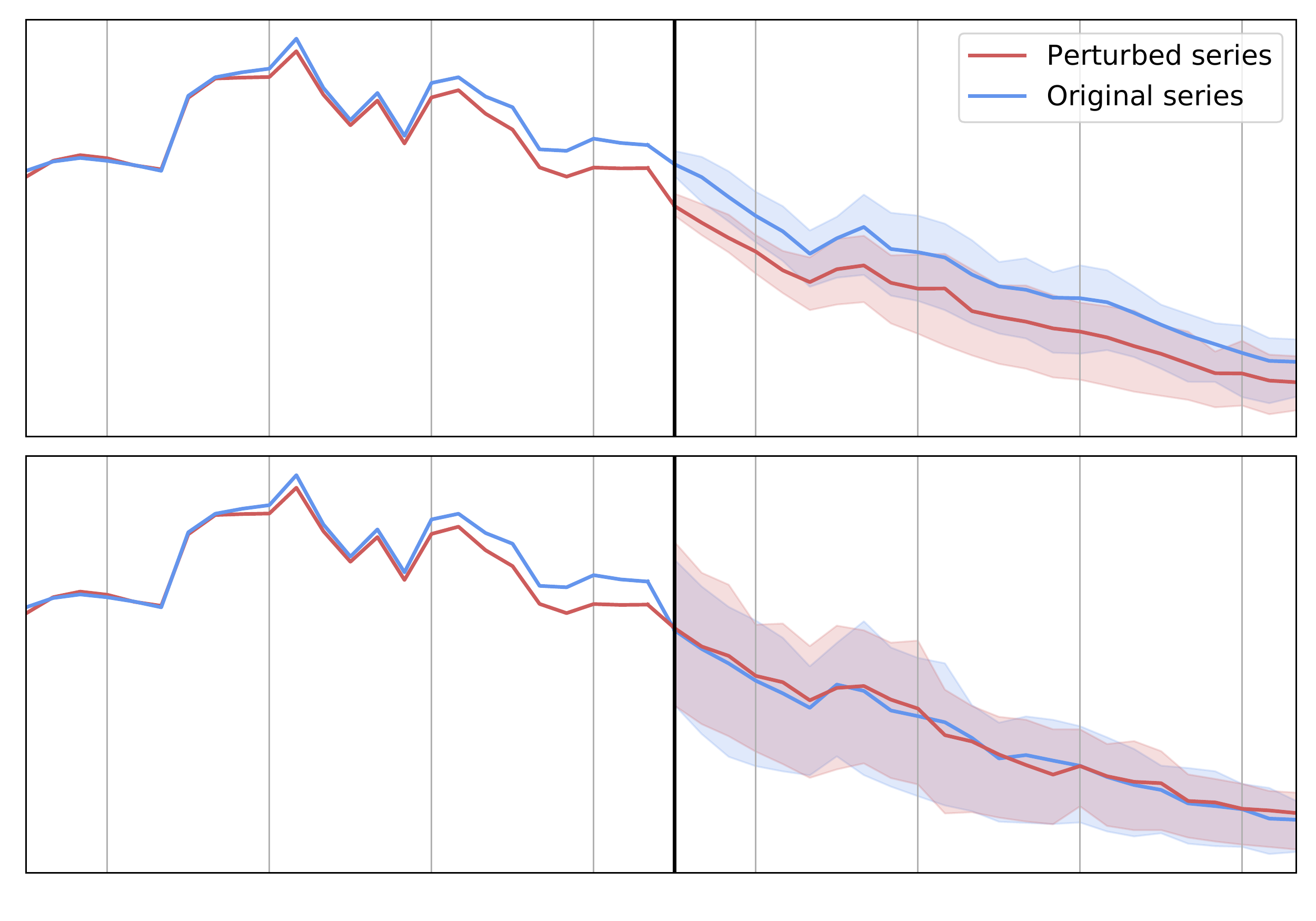}
    \caption{Predictions from vanilla DeepAR model (top) and its version with randomized smoothing we propose (bottom). Forecasts are separated from input series via the black vertical line.
    The smoothed model is more robust in the sense that its probabilistic outputs are less affected by adversarial input perturbation (which perturbs the blue series to red).}
    \label{fig:my_label}
\end{figure}

In the classical time series literature where statistical methods were predominant, studies on robust forecasting were mainly focused on model stability against outliers \citep{connor1994recurrent, gelper2010robust}.
More recently, deep learning models increasingly gained popularity and has gradually become the main workhorse of scalable forecasting~\citep{li2019enhancing,oreshkin2019n,sen2019think,fan2019multi,chen2020probabilistic,arik2020interpretable,zhou2021informer}. A distinct concept of adversarial robustness has emerged as an issue because deep neural networks are notoriously prone to small adversarial input perturbations \citep{szegedy2013intriguing, goodfellow2015explaining}.
Particularly in the context of forecasting, in \cite{dang2020adversarial}, the authors showed that deep autoregressive time series forecasting models with probabilistic outputs could suffer from such attacks.

Despite its importance, to the best of our knowledge, there is no prior work which has developed the formal concept of adversarial robustness for probabilistic forecasting models. This leaves the forecasters under the threat of adversarial attacks, endangering the decision making of mission-critical tasks.
Furthermore, the time series data possess a unique characteristic (i.e.\ the time dimension), which allows for robustness notions that are fundamentally different from $l_p$-adversarial robustness, e.g., forecasts' stability under temporal window shift (translation) or the presence of statistical outliers.
Therefore, it is necessary to establish a notion of robustness for probabilistic time series forecasting models that is general enough to encompass diverse classes of input changes and the corresponding practical requirements.

\paragraph{Contributions.} 
In this paper, we develop a framework of robust probabilistic time series forecasting, handling both theory and practice. 
To this aim, we first generalize the concept of adversarial perturbations in terms of abstract input and output transformations and provide a formal mathematical notion of robustness in terms of distributional stability of outputs when inputs are perturbed.
With these foundations established, we propose randomized smoothing for probabilistic forecasters, which enjoys theoretical robustness guarantees for distinct classes of input perturbations and potentially improves a baseline model's robustness without requiring separate learning procedures.
To establish even more pragmatic robust methods, we combine the smoothing techniques with randomized training, i.e., augmentation of training data with random noises, which is generally known to be effective in generating robust base learners.
Finally, we empirically verify that the randomizing procedures are indeed effective in rendering robust probabilistic forecasting models through extensive experiments on multiple real datasets.

\subsection{Related Work}
\label{section:related_work}

\paragraph{Classically robust forecasting via statistical methods.}
Earlier ideas on robust forecasting have mostly focused on adapting the classical techniques to deal with outliers, missing data, or change points.
A plethora of works have been developed in multiple directions, including robust versions of exponential and Holt--Winters smoothing \citep{cipra1992robust, cipra1995holt, gelper2010robust}, outlier-robust Kalman filters \citep{cipra1997kalman, ting2007kalman, agamennoni2011outlier, chang2014robust}, and statistical frameworks based on detection or filtering of anomalies \citep{connor1994recurrent, garnett2009sequential, ristanoski2013time, anava2015online, guo2016robust}.

\paragraph{Deep learning for time series forecasting.} Neural network has been applied to time series forecasting for more than half a century, and the earliest work dates back to 1960s \citep{hu1964adaptive}.
Despite the early start, neural networks found little success in the time series forecasting literature until recently.
With the explosive production of time series data and advances in neural architectures, deep learning has become increasingly popular. 
The strong performance of the deep forecasting models is especially prominent in the fields where a large collection of time series is available, such as demand forecasting in large retailers. 
Among the deep forecasting models, more relevant to the present work are the approaches that generate probabilistic forecasts. 
This is typically achieved by two avenues. 
The first approach, including~\cite{salinas2020deepar,salinas2019high,de2020normalizing,rangapuram2018deep,wang2019deep}, uses neural networks as backbone sequential model and the last layer is customized via a likelihood function. 
An alternative approach directly generates the desired quantile forecasts; see for example~\cite{wen2017multi, lim2021temporal, park2021learning, eisenach2022mqtransformer}.
Other classes of works include variance reduced training \citep{pmlr-v139-lu21d} or domain adaptaton based techniques \citep{jin2022domain}.
For comprehensive study of deep forecasting models, we refer interested readers to~\cite{benidis2020neural,hewamalage2021recurrent,alexandrov2020gluonts}. 

\paragraph{Adversarial attacks and time series.}
The seminal work of \cite{szegedy2013intriguing} demonstrated that image classification models based on deep neural networks, in spite of their high test accuracy, tend to be susceptible to hardly human-perceptible changes, which cause them to completely misclassify the inputs.
This inspired a number of works to further study effective attack schemes \citep{goodfellow2015explaining, madry2018towards, papernot2017practical, athalye2018obfuscated}.
In the time series domain, earlier works \citep{fawaz2019adversarial, karim2020adversarial} mainly focused on attacking time series classification models, and attack against probabilistic forecasting models was first devised by \cite{dang2020adversarial} using reparametrization tricks.

\paragraph{Certified adversarial defenses.}
While the adversarial training \citep{kurakin2016adversarial, madry2018towards} has been arguably the most successful defense scheme against adversarial attacks on the empirical side, its theoretical performance guarantee over perturbed data has not been established.
Towards developing defense scheme with certificates, a series of works \citep{dvijotham2018training, wong2018provable, wong2018scaling, raghunathan2018semidefinite, raghunathan2018certified, gowal2018effectiveness} suggested to directly control the local Lipschitz constant of feedforward neural networks, but such approaches were constrained to moderate-sized models.
Another line of works \citep{lecuyer2019certified, cohen2019certified, salman2019provably, li2018certified} studied randomized smoothing as a more scalable and model-agnostic approach, and successfully provided practical accuracy guarantees on classification problems up to the ImageNet scale under adversarial perturbations.
Randomized smoothing have also had applications in regression problems in the context of certifiably robust object detection \citep{chiang2020detection}.
However, we are not aware of any prior works along this direction which have considered adversarial defenses for models with probabilistic outputs, which is the standard framework for time series forecasting.

\paragraph{Exposure bias and translation robustness.}
An autoregressive (or conditional) sequence generation model may behave significantly differently in training and inference stages because in the test time, it generates outputs based on its own previous outputs, whose distribution may deviate from the ground-truth and the resulting error can be propagated \citep{bengio2015scheduled, bowman2015generating, ranzato2015sequence}.
This phenomenon, often referred to as exposure bias, has been studied and empirically addressed by a number of prior works on language models \citep{norouzi2016reward, schmidt2019generalization} and more recently on time series forecasting \citep{SANGIORGIO2020110045}.
These works are partially related to our translation robustness framework where we control propagation of errors caused by appending noisy or adversarial observations, but our approach is fundamentally distinct in that we focus on worst-case perturbations (rather than regarding the data distributions) and theoretically guaranteed solutions (rather than empirical remedies).

\section{PRELIMINARIES}
\label{section:formulation}

\subsection{Probabilistic Time Series Forecasting}

Suppose we are given a dataset of $N$ time series, where the $i$-th time series consist of observation $x_{i,t} \in \mathbb{R}$ with (optional) input covariates $z_{i,t} \in \mathbb{R}^d$ at time $t$. We drop the time series index whenever the context is clear. Examples of the input covariates include price and promotion at a certain time with the observations being the sales. 
For each time series, we observe $T$ past targets $\bx = x_{1:T} \in \cX = \bigcup_{T=1}^\infty \reals^T$ 
and all covariates $z_{1:T+\tau} \in \cZ$
to predict $\tau$ future targets $x_{T+1:T+\tau} \in \cY = \reals^\tau$. We denote a global\footnote{Forecast models are the same across all $N$ time series.} probabilistic forecaster $f:\cX \times \cZ \rightarrow \cP(\cY)$ where $ \cP(\cY)$ denotes the probability distribution on the prediction space $\cY$. With a slight abuse of notation, we describe the forecaster as
\[
(Y_{1}, \ldots, Y_{\tau}) = f(x_{1}, \ldots, x_T, z_1\ldots, z_{T+\tau}),
\]
where $(Y_{1}, \ldots, Y_{\tau})$ are random variables associated with future targets $(x_{T+1},\ldots, x_{T+\tau})$ and their full distributions are specified in terms of quantiles~\citep{wen2017multi, park2021learning} or parametric forms~\citep{salinas2020deepar}, e.g., Gaussian, Student's $t$, or negative binomial. For notational simplicity, we omit the covariates $z_{1:T+\tau}$ and concisely write
\begin{equation}
    \label{eq:prob_forecaster}
    \bY = (Y_1, \dots, Y_\tau) = f(\bx).
\end{equation}
We respectively denote as $x_{T+1}, \ldots, x_{T+\tau}$ the ground-truth future targets and $\by = (y_{1}, \ldots, y_{\tau}) = (\hat x_{T+1}, \ldots, \hat x_{T+\tau})$ the sampled prediction (or realization) of future targets from probabilistic forecaster $f$.
Since probabilistic forecast $f(\bx)$ essentially carries distributional information, we abuse notation to allow sampling from it: $(\hat x_{T+1}, \ldots, \hat x_{T+\tau})\sim f(\bx)$.

\subsection{Adversarial Attacks on Probabilistic Autoregressive Forecasting Models}
\label{subsection:adversarial_attacks}
In the probabilistic forecasting setting, the adversarial perturbation (or attack) $\bdelta$ on the input $\bx$ with given adversarial target values $\mathbf{t}_\mathrm{adv} \in \reals^m$ and a statistic\footnote{\cite{dang2020adversarial} limits $\chi$ and $\mathbf{t}_\mathrm{adv}$ to scalar ones ($m=1$).} $\chi:\reals^\tau\rightarrow \reals^m$  can be found by minimizing
\begin{align}
\label{eqn:adv_attack_dang}
    \underset{\bdelta: \|\bdelta\| \le \eta}{\argmin} \left\|
    \mathbb E_{f (\by | \bx + \bdelta)} [\chi(Y_1, \dots, Y_\tau)] - \mathbf{t}_{\textrm{adv}}
    \right\|_2^2
\end{align}
where $\eta\geq 0$ is the attack threshold, and the norm $\|\bdelta\|$ is chosen depending on the context.
The expectation $\bbE_{f(\by | \bx + \bdelta)}$ is taken over the randomness in $(Y_1, \dots, Y_\tau) = f(\bx + \bdelta)$, the output of the probabilistic forecaster \eqref{eq:prob_forecaster} on the input $\bx + \bdelta$.
The target value $\mathbf{t}_{\textrm{adv}}$ is chosen to be significantly different from $\bbE_{f(\by | \bx)}[\chi(Y_1,\dots,Y_\tau)]$.
For the case of stock price predictions, the choice of $\chi$ may be varied to express financial quantities such as buy- or sell-option prices; see \cite{dang2020adversarial} for details.

For practical experiments, we focus on attacking subsets of prediction outputs. In other words, we consider statistics of the form
\begin{align}
\chi_H(Y_1,\ldots, Y_\tau) = (Y_{h_1},\ldots, Y_{h_{m}})
\end{align}
in \eqref{eqn:adv_attack_dang}, where $H$ is a subset of prediction indices with size $m$, i.e., $H := \{h_1,\ldots, h_m \} \subset \{1,\dots,\tau\}$.
In this case, the adversary searches for a minimal norm perturbation $\bx' = \bx + \bdelta$ for which the subset of perturbed forecasts is significantly different from the original forecasts at corresponding indices, and potentially from the ground-truth values as well.

Constrained optimization \eqref{eqn:adv_attack_dang} can be relaxed into a regularized optimization problem as follows:
\begin{align}
\label{eqn:dang_adv_attack_multi_idx}
    \min_{\bdelta} L(\bdelta) := \|\bdelta\|^2 + \lambda \cdot \left\| \bbE_{f (\by | \bx + \bdelta)} [Y_H] - \boldsymbol{t}_{\textrm{adv}} \right\|_2^2
\end{align}
where $\lambda > 0$ is a hyperparameter.
The derivative of the relaxed objective \eqref{eqn:dang_adv_attack_multi_idx} with respect to $\bdelta$ could be computed via the reparametrization trick\footnote{The forecaster should support sampling from a distribution with location and/or scale parameters, which is the case for autoregressive forecasting models.} as in \cite{dang2020adversarial}, which allows us to solve the regularized problem using any first-order optimizer. 

\section{DEFINING ROBUSTNESS FOR PROBABILISTIC TIME SERIES FORECASTING}
The adversarial attack in time series forecasting has been proposed only in terms of additive input perturbation with respect to $l_p$ norm as in~\eqref{eqn:adv_attack_dang}.
However, the unique properties of time series data including the existence of time dimension, periodicity, or seasonality potentially allow for a number of distinct types of perturbation.
In this section, we generalize the notion of adversarial input perturbations in probabilistic forecasting, in order to incorporate distinct classes of input changes.
Then we define the corresponding notion of robustness which properly quantifies model sensitivity to those input perturbations. 

\subsection{Generalized Input Perturbations in Time Series Forecasting}
We consider abstract input perturbation $T_{\cX}\colon \cX \to \cX$ and output transformation $T_{\cY}\colon \cY \to \cY$. 
Given a probabilistic forecaster $f$ and an input $\bx \in \cX$, we describe the forecast output from $f$ on $\bx$ under the input perturbation $T_{\cX}$ as $f(T_{\cX}(\bx))$,
and the original forecast output with the output transformation $T_{\cY}$ applied as $\left(T_{\cY}\right)_\# f(\bx)$.
Ultimately, we want to have
\begin{align}
\label{eqn:robustness_def_intuitive}
    f \circ T_{\cX} \approx T_{\cY} \circ f
\end{align}
in probabilistic sense, toward achieving robust forecasters.
Before formally providing the detailed concept of robustness (in Section \ref{sec:robustness_definition}), we first walk through two example classes of perturbations: additive adversarial attack and time shift with new noisy observations,
and demonstrate how the transformations $T_{\cX}$ and $T_{\cY}$ can be specified. 

\subsubsection{Additive Adversarial Perturbation}
\label{ex:perturbation_robustness}
Consider the additive advsersarial perturbation which deceives the forecaster to deviate from its original forecasts on the subset $H$ of prediction indices.
We model the corresponding $T_{\cX}$ as 
\begin{align} 
    T_{\cX}(\bx) = \bx + \bdelta^\star (\bx), 
\label{eqn:adversarial_perturbation}
\end{align}
where
\begin{align}
    \bdelta^\star (\bx) = \underset{\substack{\|\bdelta\| \le \eta}}{\argmax} \left\| \bbE_{f(\by|\bx +\bdelta)} [\bY_H] - \bbE_{f(\by|\bx)} [\bY_H] \right\|^2.
\label{eqn:adversarial_vector_argmax}
\end{align}
Simply taking the output transformation as the identity map, i.e., $T_{\cY} = \mathrm{Id}$, the requirement $f\circ T_{\cX} \approx T_{\cY} \circ f$ reduces to
\begin{align*}
    f(\bx + \bdelta^\star(\bx)) \approx f(\bx).
\end{align*}
That is, we want our forecaster to be insensitive to adversarially constructed additive noise.
If $f$ is both robust in this sense and also has good prediction performance over the clean (unattacked) test dataset, its forecast quality will be retained even when it is given with adversarially perturbed time series data.

\subsubsection{Time Shift with New Noisy Observations}
\label{ex:translation_invariance}
Consider the scenario where one initially has an input series $\bx = (x_1, \dots, x_T)$, and later a set of $k \ll \tau$ new observations $\{\Tilde{x}_{T+1}, \dots, \Tilde{x}_{T+k}\}$ arrives and the entire prediction task is shifted by $k$ time steps.
Suppose that we want the initial forecasts
\begin{align*}
    f(\bx) = (Y_1, Y_2, \dots, Y_{k+1}, Y_{k+2}, \dots )    
\end{align*}
to be consistent with the forecasts
\begin{align*}
    f(\bx;\Tilde{x}_{T+1},\dots,\Tilde{x}_{T+k}) = (Y_{k+1}', Y_{k+2}', \dots )    
\end{align*}
produced after given the new observations.
In this case, we let $T_{\cX}(\bx) = (\bx; \Tilde{x}_{T+1},\dots,\Tilde{x}_{T+k})$ be the augmentation by new observations, and
\begin{align*}
    T_{\cY}(y_1, y_2, \dots, y_{k+1}, y_{k+2}, \dots) = (y_{k+1}, y_{k+2}, \dots)
\end{align*}
be the left translation operator by $k$ time steps.
Then $f\circ T_{\cX} \approx T_{\cY} \circ f$ requires that
\begin{align*}
    Y_{k+1} \approx Y_{k+1}', \, Y_{k+2} \approx Y_{k+2}', \, \dots,
\end{align*}
i.e., the two sets of forecasts should be consistent.

Note that $\Tilde{x}$'s are inherited in $T_{\cX}$, and when some of $\Tilde{x}$'s are highly anomalous or have been manipulated by an adversary, $T_{\cX}$ may largely impact the original forecasts.
In our experiments, we consider the specific case of appending a single adversarial observation
\begin{align}
    \Tilde{x}_{T+1} := (1+\rho) x_{T+1},
\label{eqn:noisy_observation}
\end{align}
which is (de-)amplified relative to the ground truth according to the adversarial parameter $\rho > -1$.

\subsection{Formal Mathematical Definition of Robustness}
\label{sec:robustness_definition}
Given a probabilistic forecaster $f$ and a transformation pair $T_{\cX}, T_{\cY}$ suitable for time series setting,
both $f(T_{\cX}(\bx))$ and $ \left(T_{\cY}\right)_\# f(\bx)$ are random variables in $\reals^\tau$ whose distributions are specified.
Let us denote,
for each prediction time point $t=1,\dots,\tau$, the associated marginal distributions by $\mu_t$ and $\mu_t'$ respectively, i.e., $ \left(f(T_{\cX}(\bx))\right)_t \sim \mu_t $ and $(\left(T_{\cY}\right)_\# f(\bx))_t \sim \mu_t'$.
We aim to formally quantify the informal notion \eqref{eqn:robustness_def_intuitive} in terms of certain metric between these distributions.

As a final ingredient for establishing the precise definition, we define $d(\bx; T_{\cX})$, which is a measure of how significant the change due to the transformation $T_{\cX}$ is, or in other words, a dissimilarity measure between $\bx$ and $T_{\cX}(\bx)$. 
For the case of additive adversarial perturbation in Section~\ref{ex:perturbation_robustness}, a natural choice for $d$ would be $d(\bx; T_{\cX}) = \|\bdelta^\star (\bx)\|$.
For the time-shift setup in Section~\ref{ex:translation_invariance}, we take $d$ of the form
\begin{align*}
    d(\bx; T_{\cX}) = D \left( \Tilde{\bx}_{T+1:T+k}; \hat{\bx}_{T+1:T+k} \right)
\end{align*}
where $\hat{x}_{T+1:T+k}$ are the model's initial point forecasts on the first $k$ future time points, and $D \ge 0$ satisfies $D(\by'; \by) = 0$ iff $\by' = \by$.
We take $D(\by';\by) = \|\by'-\by\|_2$ in this paper for the sake of establishing theoretical guarantees, but $D$ can also be chosen in ways that involve likelihood to less penalize large deviations, e.g., $D(\by'; \by) = -\log \left( q(\by') \big/ q(\by) \right)$, where $q$ is the joint density function for the model's $k$-step ahead predictions.

\begin{definition}
\label{def:robustness}
Let $f$ be a probabilistic forecaster,  $\bx \in \cX$, and $T_{\cX}, T_{\cY}$ the input, output transformations with marginal distributions $\mu_t, \mu_t'$, respectively. 
Then, $f$ is $\varepsilon$--$\eta$ robust at $\bx$ with respect to the transformation pair $(T_{\cX}, T_{\cY})$ if, provided that $d (\bx; T_{\cX}) < \eta$, for any $t=1,\dots,\tau$, we have
\begin{align}
\label{eqn:robustness}
    W_1 \left( \mu_t, \mu_t' \right) < \varepsilon.
\end{align}
\end{definition}

\paragraph{Connection to adversarial attacks.} 
In the additive adversarial attack we detailed in Section \ref{subsection:adversarial_attacks}, it is assumed that $\chi$ is chosen by the adversary.
Therefore, a defense scheme against the attack \eqref{eqn:adv_attack_dang} would naturally involve minimizing an objective of the form
\begin{align}
\label{eqn:adv_defense_dual}
    \sup_{\|\bdelta\| \le \eta} \sup_{\chi \in \mathcal{F}} \left( \bbE [\chi(f(\bx + \bdelta))] - \bbE [\chi(f(\bx))] \right),
\end{align}
where $\mathcal{F}$ denotes the collection of $\chi$'s which the adversary could choose from.
Note that if $\|\cdot\| = \|\cdot\|_2$, and if $\cF$ consists of $L$-Lipschitz continuous functions for some $L>0$, the inner maximization in \eqref{eqn:adv_defense_dual} results in a constant multiple of 1-Wasserstein (or $W_1$) distance between the distributions of $f(\bx + \bdelta)$ and $f(\bx)$, due to Kantorovich-Rubinstein duality.
This interpretation motivates our choice of $1$-Wasserstein distance as the measure of local distributional change in the robustness definition \eqref{eqn:robustness}.

\paragraph{Connection to quantile forecasts.}
Here we provide another, more general perspective on the reason for formulating Definition~\ref{def:robustness} in terms of $W_1$ distance.
Probabilistic forecasts are often characterized via quantiles; MQ-RNN \citep{wen2017multi} directly performs quantile regression, and for sampling-based forecasters such as DeepAR \citep{salinas2020deepar}, sample quantiles are used to compute the prediction intervals.
This practice of using quantiles as important quantities is fortuitously aligned with the following interpretation of $W_1$ distance as the average quantile difference;
if $F, G$ are respectively the cumulative distribution function of a real-valued random variable and $\mu, \nu$ are the corresponding probability distributions, then
\begin{align}\label{eq:w_1_quantile}
    W_1 (\mu, \nu) = \int_0^1 |F^{-1} (u) - G^{-1}(u)| \, du .
\end{align}
That is, our robustness definition requires that a model's quantile estimates, in the average sense over probability levels, should not be significantly affected by small input perturbations with respect to $d_{\cX}$.

\section{THEORY AND FRAMEWORKS}
\label{section:theory}
In this section, we develop methodologies for robust forecasting based on randomized smoothing, covering the two classes of adversarial perturbations we considered in Section~\ref{sec:robustness_definition}. 
We establish and discuss the theoretical robustness guarantees of these smoothing-based methods.
Additionally, we revisit the randomized training (data augmentation with noises) widely adopted by practitioners as a strategy for enhancing base forecasters for randomized smoothing techniques.

\subsection{Randomized Smoothing for Guaranteed Robustness Against $l_2$ Perturbations}
\label{subsection:randomized_smoothing}
We first develop a robustness framework using randomized smoothing for additive adversarial perturbations (covered in Section~\ref{ex:perturbation_robustness}) with $d(\bx; T_{\cX}) = \|\bdelta\|_2$.

\paragraph{One-step ahead predictors.}
Consider the simple setting where $f$ is a random function from $\reals^T$ to $\reals$, which we will extend to multivariate cases later.
Let us denote by $F_{\bx}$ the cumulative distribution function (cdf) of the random variable $f(\bx)$, i.e., 
\begin{align*}
    F_{\bx} (r) := \prob[f(\bx)\le r]
\end{align*}
for $r\in \reals$.
Given the smoothing paramter $\sigma > 0$, we define the smoothed version $g_\sigma$ of $f$ as the random function from $\reals^T$ to $\reals$ with cdf
\begin{align}
\label{eqn:dist_mix}
    G_{\bx, \sigma} (r) &= \prob [g_\sigma(\bx) \le r] \\
    &= \underset{\bz \sim \cN(0,\sigma^2 I)}{\bbE} [F_{\bx + \bz}(r)] = \int F_{\bx+\bz}(r) p_\sigma (\bz) \, d\bz, \nonumber
\end{align}
where $p_\sigma (\bz)$ is the density function of the multivariate Gaussian distribution $\cN(0,\sigma^2 I)$.
Below, we provide a theoretical guarantee on the robustness of the smoothed probabilistic predictor $g_\sigma$, analogous to prior results \citep{cohen2019certified, salman2019provably, chiang2020detection} for deterministic setups.

\begin{theorem}
\label{thm:smoothing}
Let $f$ be a random function from $\reals^T$ to $\reals$, and let $g_\sigma$ be as in \eqref{eqn:dist_mix}.
Given $\bx \in \reals^T$,
we have the inequality
\begin{align}
\label{eqn:thm_bound}
\begin{aligned}
    \mathrm{Ro}(\bx; \sigma) & := \limsup_{\|\bdelta\|_2 \to 0} \frac{W_1(G_{\bx,\sigma}, G_{\bx+\bdelta,\sigma})}{\|\bdelta\|_2} \\
    & \le \frac{1}{\sigma} \int_{-\infty}^{\infty} \phi\left(\Phi^{-1}(G_{\bx,\sigma}(r))\right) dr 
\end{aligned}
\end{align}
provided that the integral on the right hand side is locally bounded at $\bx$,
where $\phi, \Phi$ denote the pdf and cdf of the standard normal distribution.
\end{theorem}

Note that if $\mathrm{Ro}(\bx;\sigma) < \infty$, then $f$ is $O(\eta)$--$\eta$ robust for $\eta$ small enough, in the sense of Definition~\ref{def:robustness} with respect to adversarial perturbations, in the one-step ahead prediction case.

\RestyleAlgo{ruled}

\begin{algorithm}[t]
\caption{Randomized smoothing for probabilistic forecasters}\label{alg:randomized_smoothing}
\KwInput{Multi-horizon sample-based forecaster $f$, Input series $\bx = (x_1,\dots,x_T)$, $\tau$, $n$, $\sigma^2$}
\KwOutput{$n$ sample paths $\hat{\bx}_{T+1:T+\tau}^{(j)}$ from $g_\sigma$ ($j=1,\dots,n$)}
\For{$j = 1, \dots, n$}{
    $\zeta_1, \dots, \zeta_T \sim \cN(0,\sigma^2)$ i.i.d.\\
    $\Tilde{\bx} \gets (x_1 + \zeta_1, \dots, x_T + \zeta_T)$\\
    $\hat{\bx}_{T+1:T+\tau}^{(j)} \sim f(\Tilde{\bx})$
}
\end{algorithm}

\paragraph{Finiteness of $\mathrm{Ro}(\bx;\sigma)$.}
Provided that the cdf $G_{\bx,\sigma}(r)$ of the smoothed random variable $g_\sigma (\bx)$ has nonzero derivative at all $r\in \reals$, we can make the change of variable $u=G_{\bx,\sigma}(r)$ to rewrite \eqref{eqn:thm_bound} as
\begin{align*}
    \mathrm{Ro}(\bx;\sigma) \le \frac{1}{\sigma} \int_{0}^1 \phi(\Phi^{-1}(u)) \left(G_{\bx,\sigma}^{-1}\right)'(u) \, du .
\end{align*}
Note that the quantity $\left(G_{\bx,\sigma}^{-1}\right)'(u)$ is the quantile density function of $g_\sigma(\bx)$, while $\phi\left(\Phi^{-1}(u)\right)$ is the inverse quantile density of the standard Gaussian.
Therefore, intuitively speaking, we expect $\mathrm{Ro}(\bx;\sigma)$ to be finite unless the quantiles of $g_\sigma (\bx)$ blow up too quickly compared to those of the normal distribution as $u\to 0, 1$.
Indeed, the following Lemmas show that $\mathrm{Ro}(\bx;\sigma) < \infty$ holds under mild, practical assumptions on $f$.

\begin{lemma}
\label{lemma:finiteness}
Let $f$ and $g_\sigma$ be as in Theorem~\ref{thm:smoothing}. Suppose that there exists a function $\varphi: \reals_+ \to \reals_+$ such that $\prob[|f(\bx)| \ge r] \le \varphi (r)$ for any $\bx \in \reals^T$ and $r > 0$, and
\begin{align}
\label{eqn:varphi_L1}
    \int_0^\infty \varphi (r) \, dr < \infty .
\end{align}
Then there exists $C > 0$, depending only on $\varphi$ and $\sigma$, such that $\mathrm{Ro}(\bx;\sigma) < C$ for all $\bx \in \reals^T$.
\end{lemma}

\begin{lemma}
\label{lemma:distributions}
Suppose that $f$ is parametrized with globally bounded mean and variance.
Then $f$ satisfies the assumptions of Lemma~\ref{lemma:finiteness}.
\end{lemma}
In particular, the most commonly used distributions including Gaussian, generalized Student's $t$ ($\nu>2$), or negative binomial are all allowed if the mean and scaling parameters are bounded. 

\paragraph{Multi-horizon predictors.}
In the case when $f$ is a random function from $\reals^T$ to $\reals^\tau$ as in the multi-horizon forecasting setup, we can apply Theorem~\ref{thm:smoothing} and Lemmas~\ref{lemma:finiteness},~\ref{lemma:distributions} to each component function of $f$.
This directly implies that $f$, under suitable assumption, is $O(\eta)$--$\eta$ robust with respect to the $l_2$ adversarial perturbation in the sense of Definition~\ref{def:robustness}, which we formally restate in the following as Corollary~\ref{cor:finiteness}.
Therefore, we take the (component-wisely) smoothed version $g_\sigma$ of the baseline predictor $f$ as our predictor with robustness guarantees, which is presented in Algorithm~\ref{alg:randomized_smoothing}.
Note that we implement $g_\sigma$ by directly sampling its output paths, which is done by sampling independently over randomness in $\bz$ and $f$. 

\begin{corollary}
\label{cor:finiteness}
Let $f$ be a probabilistic multi-horizon forecaster, denoted $f(\bx) = (f_1(\bx), \dots, f_\tau(\bx))$.
Suppose that $\varphi : \reals_+ \to \reals_+$ satisfies $\prob[|f_j(\bx)| \ge r] \le \varphi (r)$ for any $\bx \in \reals^T$, $r > 0$ and $j=1,\dots,\tau$, and \eqref{eqn:varphi_L1} holds.
Then there exists a constant $C > 0$ depending only on $\varphi$ and $\sigma$ such that for any $\eta > 0$, the smoothed forecaster $g_\sigma$ defined by applying the smoothing \eqref{eqn:dist_mix} to each component $f_j(\bx)$ is $C\eta$--$\eta$ robust at all $\bx$ with respect to $T_{\cX}$ as in \eqref{eqn:adversarial_perturbation} with $\|\cdot\| = \|\cdot\|_2$, $T_{\cY} = \mathrm{Id}$ and $d(\bx; T_{\cX}) = \|\bdelta^\star(\bx)\|$.
\end{corollary}

\paragraph{Possible generalization of the theorems.}
We point out that the domain of $f$ could be replaced with some representation space $\cW$ without altering the essence of the theorems.
If $\Psi: \cX \to \cW$ is an invertible representation map such that $d_{\cX}(\bx; T_{\cX}) = \|\Psi(T_{\cX}(\bx)) - \Psi(\bx)\|_2$, then one can simply apply Theorem \ref{thm:smoothing} and Corollary \ref{cor:finiteness} to $f\circ \Psi^{-1}$, which is a random function from $\cW$ to $\reals$.
Thus, albeit the theorems seem to only provide guarantees for local $l_2$ perturbations, one can extend the results to fundamentally distinct types of perturbations by choosing an appropriate $\cW$, e.g., the frequency domain.

\subsection{Noisy Autoregressive Inference for Robustness Under Time Shift}
\label{subsection:future_smoothing}

\begin{algorithm}[t]
\caption{Future smoothing for probabilistic autoregressive forecasters}
\label{alg:future_smoothing}
\KwInput{Probabilistic autoregressive forecaster $f$, Input series $\bx = (x_1,\dots,x_T)$, $\tau$, $n$, $\sigma^2$, (Optionally) Noisy observations $(y_1,\dots,y_k) = (\Tilde{x}_{T+1}, \dots, \Tilde{x}_{T+k})$}
\KwOutput{Sample forecasts $\hat{x}_{T+i}^{(j)}$ ($i=k+1,\dots,k+\tau$, $j=1,\dots,n$)}
\For{$i = 1, \dots, k + \tau$}{
    \eIf{$i \le k$}{
        $y_{i-1} \leftarrow \Tilde{x}_{T+i-1}$
    }
    {
        $y_{i-1} \leftarrow \hat{x}_{T+i-1}$ \\
        \For{$j = 1, \dots, n$}{
            $\zeta_1, \dots, \zeta_{i-1} \leftarrow \cN(0, \sigma^2)$ i.i.d. \\
            $\hat{x}_{T+i}^{(j)} \leftarrow f\left(\bx; y_1+\zeta_1, \dots, y_{i-1}+\zeta_{i-1}\right)$
        }
        $\hat{x}_{T+i} \leftarrow \frac{1}{n} \sum_{j=1}^n \hat{x}_{T+i}^{(j)}$
    }
}
\end{algorithm}

In this section, we propose a strategy, Algorithm~\ref{alg:future_smoothing}, for achieving robustness against the time-shift setup of Section~\ref{ex:translation_invariance}.
As randomized smoothing buffers the effect of input perturbation by averaging over noised inputs, we exploit random noises to buffer the effect of appending the noisy observation $\Tilde{x}_{T+1}$.
Note that because the uncertainty is now within future times, we perform smoothing over future time indices.

An autoregressive forecaster $f$ can be generally described in the form $f(\bx) = (Y_1, Y_2, \dots)$, where $\bx \in \reals^T$ is an input series and
\begin{align}
\label{eqn:f_autoregressive}
    Y_h = f^{(h)}\left( \bx; Y_1, \dots, Y_{h-1} \right)
\end{align}
for some random functions $f^{(h)}$ from $\reals^{T+h-1}$ to $\reals$, for $h=1,2,\dots$.
For fixed $\sigma > 0$ and $\bx\in\reals^T$, consider the smoothed version $g_\sigma^{(h)}$ of $f^{(h)}$, defined as
\begin{align*}
    & g_\sigma^{(h)} (\bx; y_1, \dots, y_{h-1}) \\
    & \quad \quad = \underset{\boldsymbol{\zeta} \sim \cN(0, \sigma^2 I_{h-1})}{\bbE} \left[ f^{(h)}(\bx; y_1 + \zeta_1, \cdots, y_{h-1} + \zeta_{h-1}) \right] 
\end{align*}
where we noise only the variables $y_1,\dots,y_{h-1}$ but not $\bx$.
Now we define $g_\sigma$, the smoothed version of $f$, by $g_\sigma(\bx) = (Y_1, Y_2, \dots)$, where
\begin{align}
\label{eqn:g_sigma_future_smoothing}
    Y_h = g_\sigma^{(h)} \left( \bx; y_1, \dots, y_{h-1} \right)
\end{align}
for $h=1,2,\dots$, and $y_j \in \reals$ ($j=1,\dots,h-1$) denotes either 1) the mean forecast from $g_\sigma$ at that time point (i.e.\ expected value of $Y_j$) if the true value is unobserved, or 2) the given ground-truth value if a new observation at that time has arrived.
Algorithm~\ref{alg:future_smoothing} presents the procedural implementation of $g_\sigma$ based on sampling.
The theoretical justification for performing smoothing in this way is Corollary~\ref{cor:future_smoothing}, which states the robustness property of the smoothed forecaster $g_\sigma$ with respect to the notions provided in Sections~\ref{ex:translation_invariance} and \ref{sec:robustness_definition}, which is done by applying the smoothing framework developed in Section~\ref{subsection:randomized_smoothing} under a slightly different setting.

\begin{corollary}
\label{cor:future_smoothing}
Let $f$ be an autoregressive forecaster defined as \eqref{eqn:f_autoregressive} and suppose that $\varphi: \reals_+ \to \reals_+$ satisfies \eqref{eqn:varphi_L1} and
\begin{align*}
    \prob\left[ \left|f^{(h)}(\bx; y_1, \dots, y_{h-1})\right| \ge r\right] \le \varphi (r)
\end{align*}
holds for each $h=1,2,\dots$.
Then, for all $\eta > 0$, the forecaster $g_\sigma$ defined as \eqref{eqn:g_sigma_future_smoothing} is $O(\eta)$--$\eta$ robust at all $\bx \in \reals^T$ with respect to $T_{\cX}, T_{\cY}$ of Section~\ref{ex:translation_invariance} and
\begin{align*}
    d(\bx; T_{\cX}) = \left\| \Tilde{\bx}_{T+1:T+k} - \hat{\bx}_{T+1:T+k} \right\|_2 ,
\end{align*}
where $\tilde{\bx}_{T+1:T+k}$ are the values appended by $T_{\cX}$ and $\hat{\bx}_{T+1:T+k}$ are the $k$-step mean predictions from the initial forecasts $g_\sigma (\bx)$.
\end{corollary}

\begin{figure*}[t]
\captionsetup[subfigure]{justification=centering}
    \centering
    \begin{subfigure}{0.33\linewidth}
    \centering
    \includegraphics[scale=0.32]{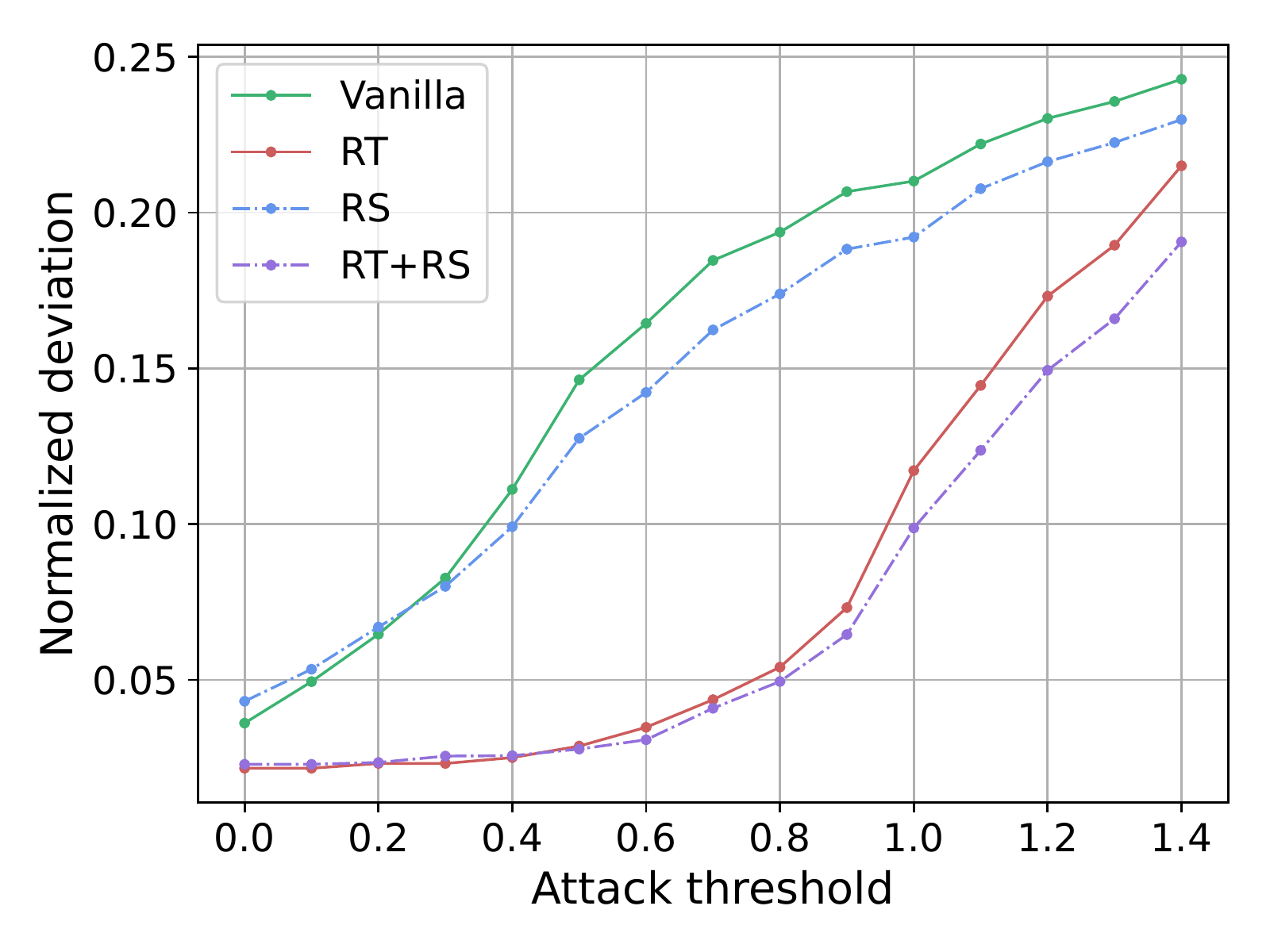}
    \caption{Exchange Rate, $\tau = 30$, $\{H\} = \{\tau\}$}
    \label{subfig:exchange_rate_tau_30_H_30}
    \end{subfigure}
    \hspace{-.2cm}
    \begin{subfigure}{0.33\linewidth}
    \centering
    \includegraphics[scale=0.32]{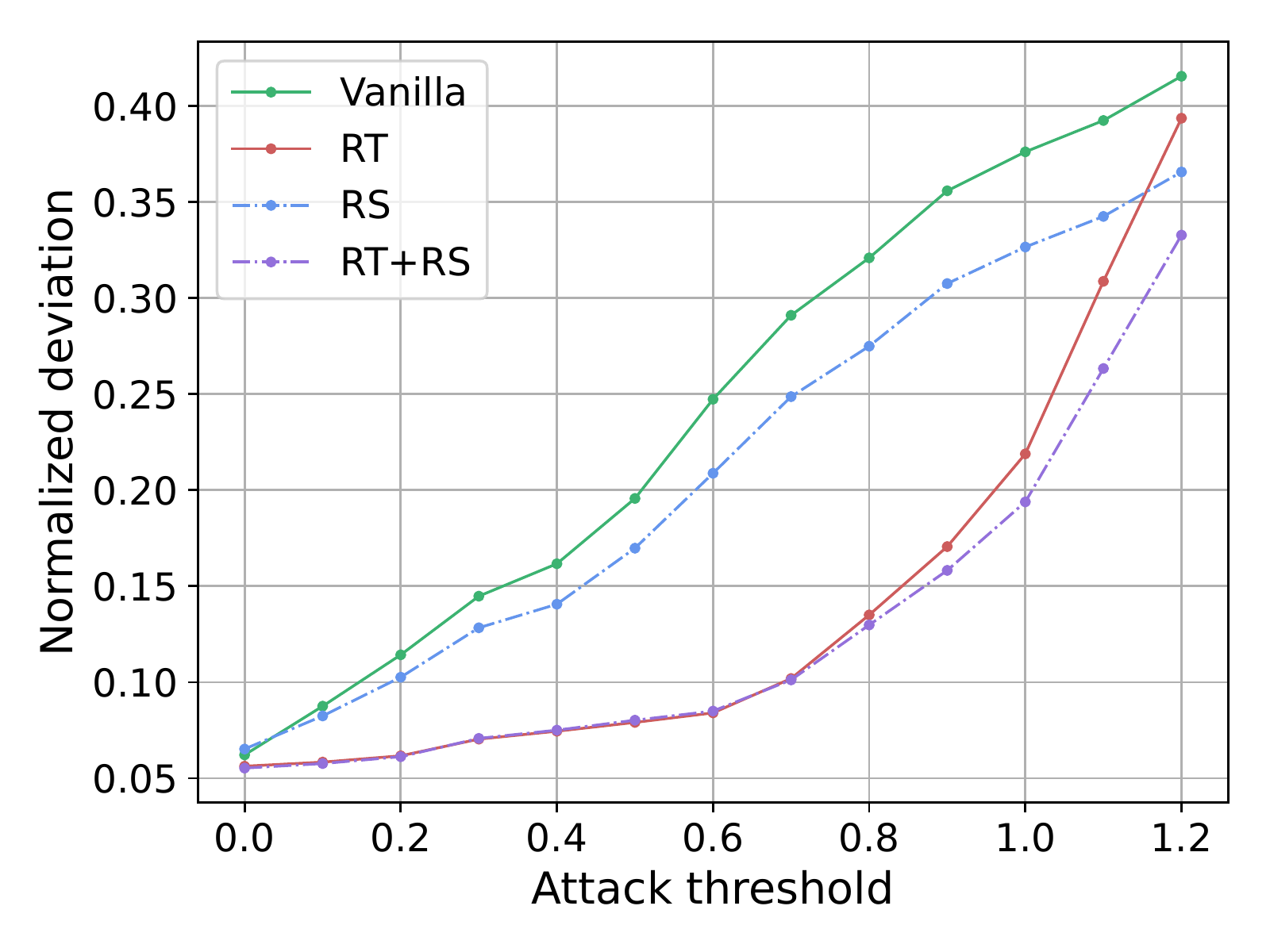}
    \caption{M4-Daily, $\tau = 14$, $\{H\} = \{\tau\}$}
    \label{subfig:m4_daily_tau_14_H_14}
    \end{subfigure}
    \hspace{-.2cm}
    \begin{subfigure}{0.33\linewidth}
    \centering
    \includegraphics[scale=0.32]{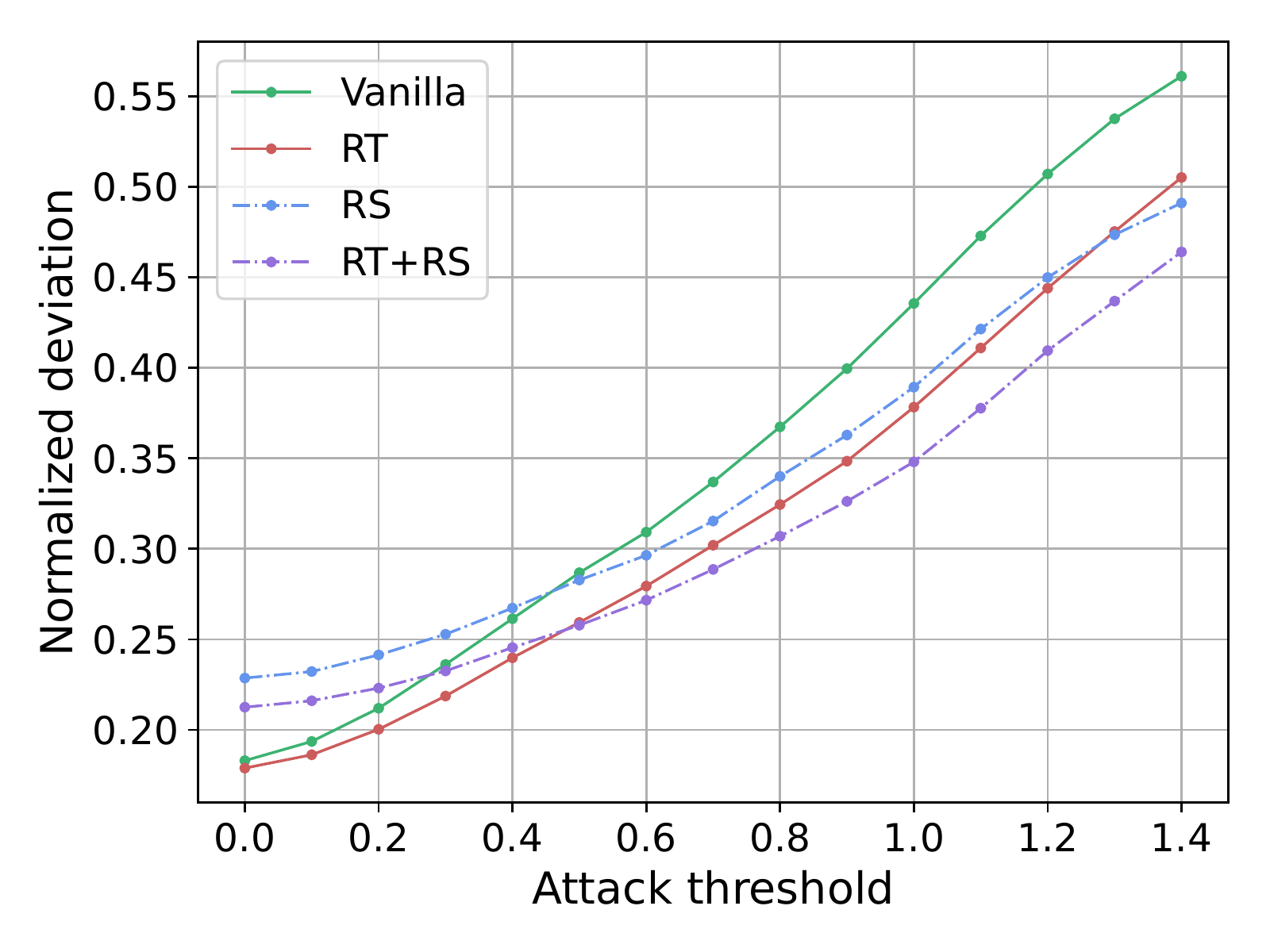}
    \caption{Traffic, $\tau = 24$, $\{H\} = \{\tau\}$}
    \label{subfig:traffic_tau_24_H_24}
    \end{subfigure}
    \caption{$\mathrm{ND}_H$ from DeepAR models on different datasets. Randomized smoothing (RS) of baseline models uses $\sigma = 0.5$. Randomized training (RT) uses $\sigma_{\mathrm{tr}} = 0.1$.}
    \label{fig:adversarial_accuracy_main}
\end{figure*}

\subsection{Training the Baseline Model via Random Data Augmentation}
\label{subsection:randomized_training}
Having a baseline model $f$ with solid prediction performance over noised inputs is the key for achieving success in randomized smoothing, which is expected to improve the theoretical certificate \eqref{eqn:thm_bound} as well as practical performance \citep{cohen2019certified}.
This encourages us to additionally apply \textit{randomized training}, which augments the training dataset with random noises to adapt the baseline forecaster to random perturbations. 
Note that randomized training and smoothing are different schemes; in particular, the former requires training from scratch while the latter does not.
In the end, we take advantage of both techniques to achieve effective robust forecasters.

\section{EXPERIMENTS}
\label{section:experiments}

In this section, we demonstrate the effects of our frameworks for the DeepAR \citep{salinas2020deepar} implementation within GluonTS \citep{alexandrov2020gluonts} on real datasets, including the M4-Daily, Exchange Rate, Traffic and UCI Electricity datasets preprocessed as in \cite{salinas2019high}.
We use DeepAR as it is the standard model with properties of being sampling-based and autoregressive, which Algorithms~\ref{alg:randomized_smoothing} and \ref{alg:future_smoothing} respectively require.

We specifically focus on how point forecasts from the model change under input transformations of Sections \ref{ex:perturbation_robustness} and \ref{ex:translation_invariance}, and quantitatively assess them using the normalized deviation (ND)
\begin{align}
\label{eqn:normalized_deviation}
    \mathrm{ND}_H = \frac{\sum_{k=1}^N \sum_{h\in H} |\hat{x}_{k,T+h} - x_{k,T+h}^{\textrm{ref}}|}{\sum_{k=1}^N \sum_{h\in H} |x_{k,T+h}^{\textrm{ref}}|}.
\end{align}
Here $H \subset \{1,\dots,\tau\}$ is the set of prediction indices of interest, $N$ is the size of the test dataset, $\hat{x}_{k,T+h}$ is the model's prediction for $(T+h)$-th time step for the $k$th (possibly transformed) test series, and $x_{k,T+h}^{\textrm{ref}}$ is the corresponding reference value (which may either be the ground-truth value or the model output before transformation, depending on the setup).

\subsection{Prediction Performance Under Additive Adversarial Attacks}
\label{subsection:experiments_adversarial_perturbation}

In this section, we consider the setup of Section~\ref{ex:perturbation_robustness}, where the input transformation corresponds to the adversarial attack of \cite{dang2020adversarial}.

\paragraph{Experimental setup.}
Following \cite{dang2020adversarial}, we choose the relative $l_2$ norm
\begin{align*}
    \|\bdelta\|_{\bx} := \left( \sum_{i=1}^t \left( \delta_i / x_i \right)^2 \right)^{1/2}
\end{align*}
as the measure of perturbation magnitude.
Given the set $H \subset \{1,\dots,\tau\}$ of attack indices and attack threshold $\eta > 0$, we solve the problem \eqref{eqn:dang_adv_attack_multi_idx} with different choices of $\mathbf{t}_{\textrm{adv}}$ and $\lambda$, and among the resulting approximate solutions $\bdelta$ meeting the norm constraint $\|\bdelta\|_{\bx} \le \eta$, we measure the largest point forecast error in terms of normalized deviation \eqref{eqn:normalized_deviation}.
Here we take the ground-truth future values as reference values, i.e., $x_{k,T+h}^{\textrm{ref}} = x_{k,T+h}$.
As the attack is performed under norm constraint with respect to $\|\cdot\|_{\bx}$, we accordingly use \textit{relative} noises in both randomized smoothing and randomized training.
That is, if the given input is $\bx = (x_1,\dots,x_T)$ and the randomizing variance is $\sigma^2$, then we use 
$(x_1(1+\zeta_1), \dots, x_t(1+\zeta_t))$ as noised inputs, where $\zeta_1, \dots, \zeta_t$ are i.i.d.\ samples from $\cN(0,\sigma^2)$.
Note that in this case, our theoretical results can be applied with locally scaled version of $f$ with respect to each input $\bx$.
To distinguish the two randomizing procedures, we respectively denote by $\sigma^2$ and $\sigma_{\textrm{tr}}^2$ the variance values of noises used in smoothing and data augmentation for training.

\begin{figure*}[ht]
\captionsetup[subfigure]{justification=centering}
    \centering
    \begin{subfigure}{0.33\linewidth}
    \centering
    \includegraphics[scale=0.34]{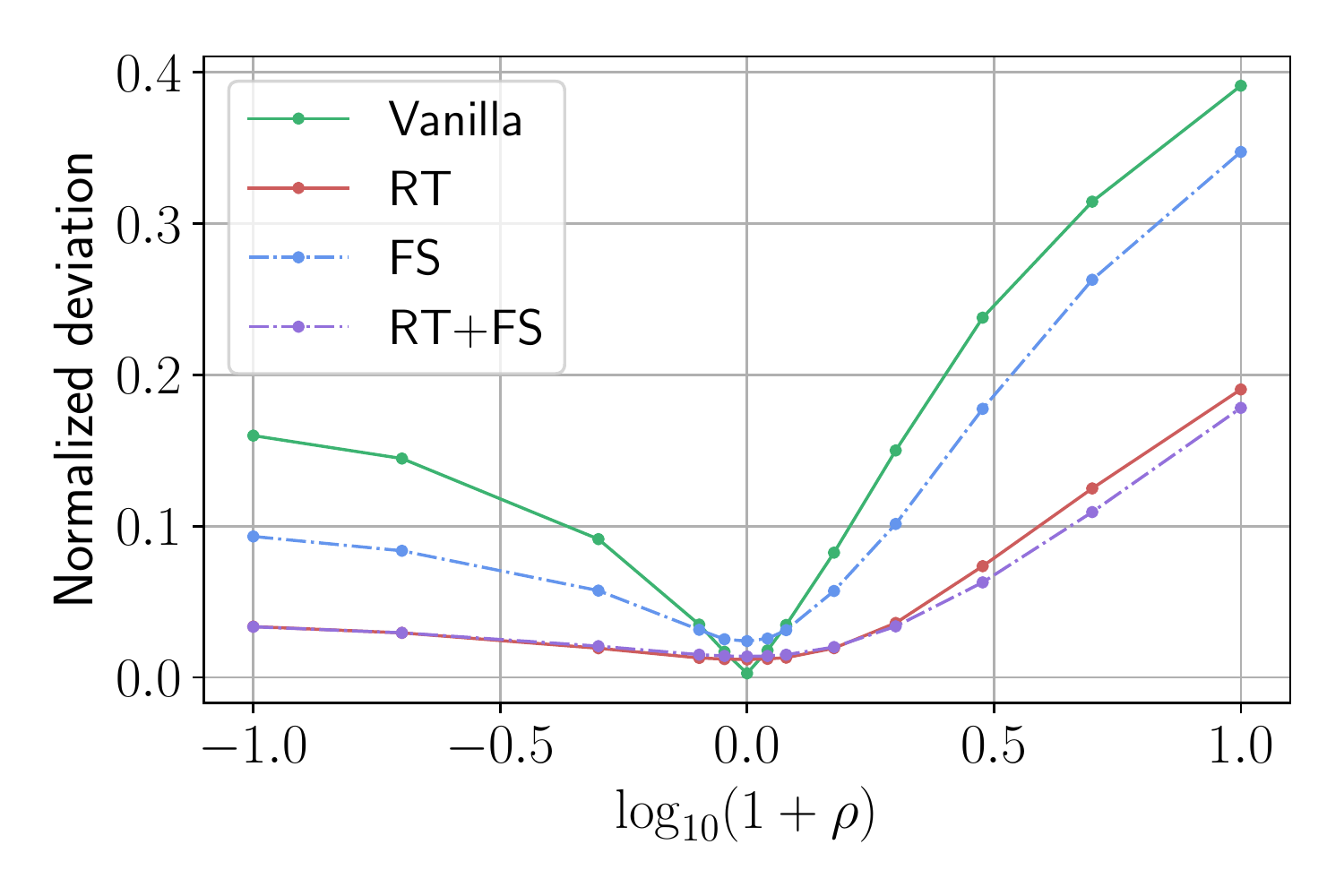}
    \caption{Exchange Rate}
    \label{subfig:translation_exchange_rate}
    \end{subfigure}
    \hspace{-.2cm}
    \begin{subfigure}{0.33\linewidth}
    \centering
    \includegraphics[scale=0.34]{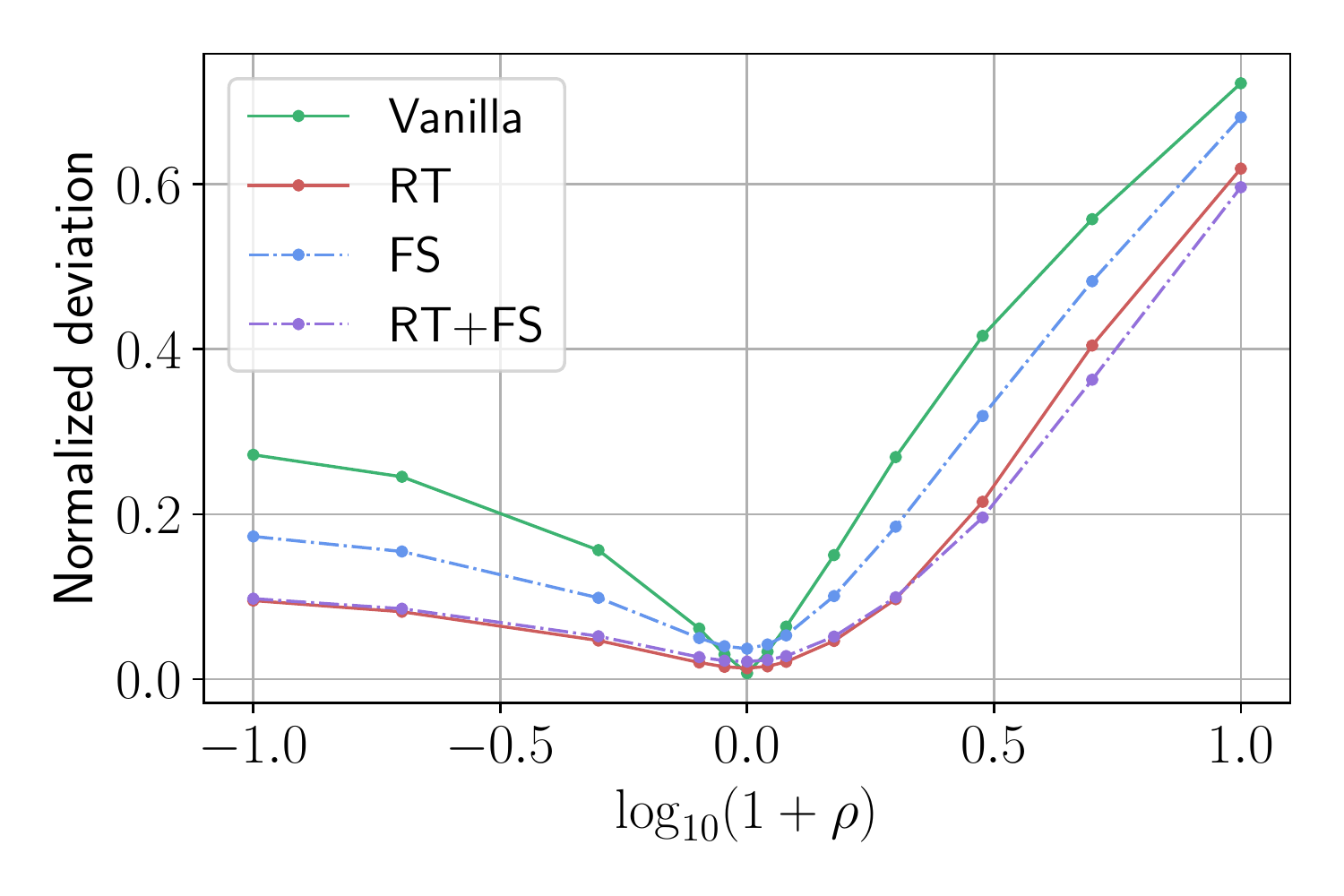}
    \caption{M4-daily}
    \label{subfig:translation_m4}
    \end{subfigure}
    \hspace{-.2cm}
    \begin{subfigure}{0.33\linewidth}
    \centering
    \includegraphics[scale=0.34]{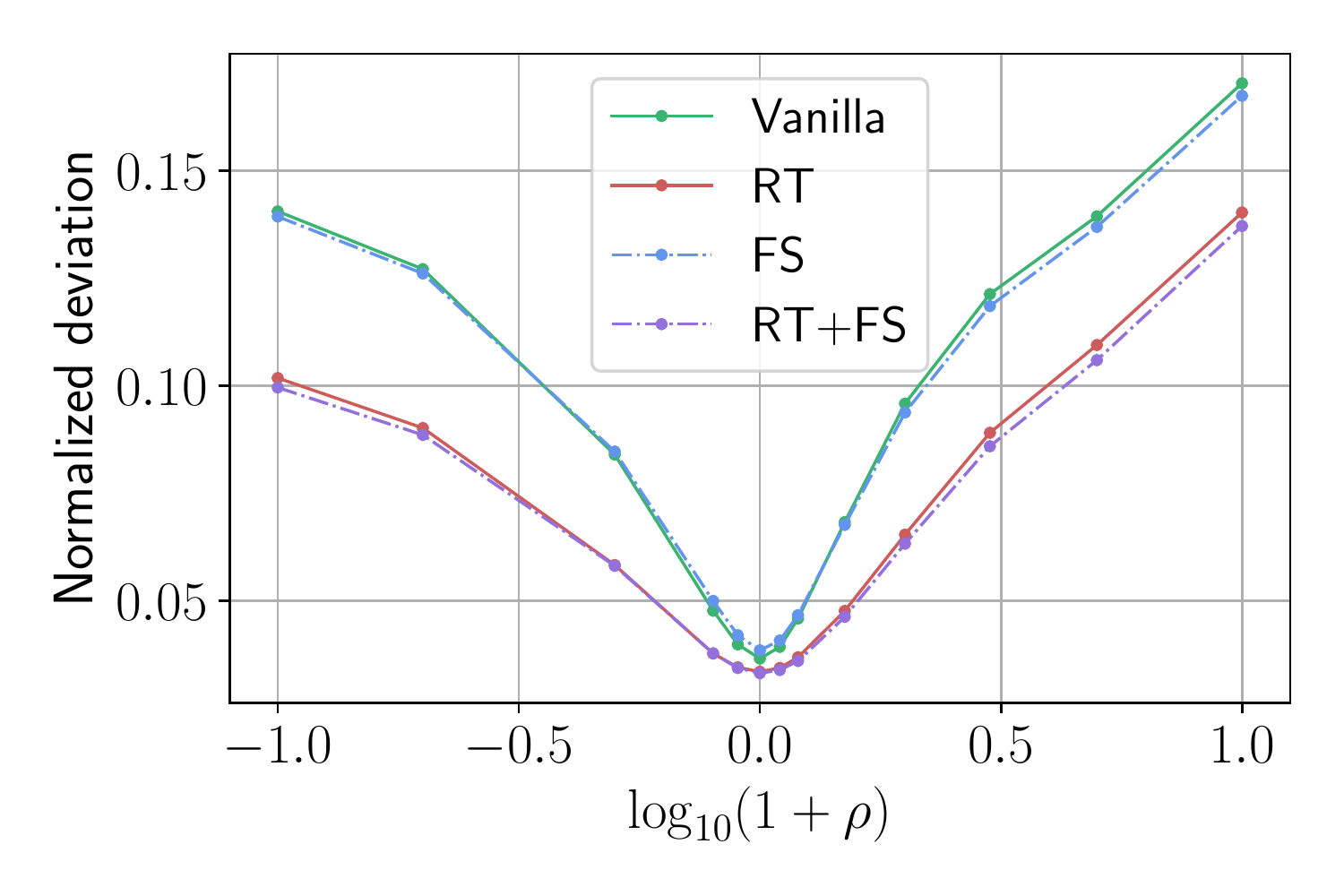}
    \caption{Traffic}
    \label{subfig:translation_traffic}
    \end{subfigure}
    \caption{$\mathrm{ND}_{\textrm{rel}}$ from DeepAR models on different datasets. Randomized training (RT) uses $\sigma_{\mathrm{tr}} = 0.1$. Future smoothing (FS) uses $\sigma = 1.0$ for the first two datasets, and $\sigma = 0.1$ for the Traffic dataset. 
    }
    \label{fig:translation_invariance_main}
\end{figure*}

\paragraph{Results and discussion.}
Figure \ref{fig:adversarial_accuracy_main} respectively compares the performance of vanilla DeepAR models (trained without data augmentation; solid green line) with their corresponding smoothed versions (labeled RS; dashed blue line), and the random-trained models (labeled RT; solid red line) with their smoothed versions (labeled RS$+$RT; dashed purple line).
We observe three important points.
First, for the majority of cases, randomized smoothing provides statistically significant improvement to vanilla models' performance under attacks with moderate to high threshold values $\eta$ (see Table~\ref{tab:adversarial_perturbation_appendix} in the appendix), at the cost of possibly slightly worsening the performance for small $\eta$.
Second, RT models tend to strongly outperform the corresponding vanilla models, uniformly over the levels of attack threshold $\eta$.
Third, further smoothing the RT model (RS$+$RT) can improve upon RT and in these cases, RS$+$RT achieves the best empirical performance against attacks with high values of $\eta$.
Table~\ref{tab:adversarial_perturbation_appendix}, provided in the appendix, displays full experiment results for all datasets and distinct attack indices.

We emphasize that only the smoothed models are the ones that come with robustness certificates, and in particular, we find that RS$+$RT is a promising methodology, supported from both theoretical and empirical sides.
Additionally, randomized smoothing is readily applicable to any pre-trained baseline model as a post-processing step without the cost of model retraining, and still offers a potential direction for improving the model's robustness.
We believe that these points constitute important practical values of the smoothing technique in general.

\paragraph{Randomized training and prediction performance.}
As an aside, we unexpectedly find that RT tends to improve the usual prediction performance of forecasting models (see Table~\ref{tab:randomizing_and_clean_accuracy} in the appendix). %
That is, randomizing the training data may positively impact a model's generalization in the time series domain, which we believe, is an interesting phenomenon in its own right.
In the appendix, we provide further discussion on this point, connecting the observation to prior works on generalization and training with noise.

\subsection{Forecast Consistency Under Time Shift with Noisy Observation}
\label{subsection:experiments_translation_invariance}

In this section, we examine the setup of Section~\ref{ex:translation_invariance} with $k=1$, where we append an adversarial observation to input series.

\paragraph{Experimental setup.}
Given a series $\bx \in \reals^T$, we append $\tilde{x}_{T+1}$ as in~\eqref{eqn:noisy_observation} with various values of $\rho$ within the range $-1 \le \log_{10}(1+\rho) \le 1$.
As a metric, we compute the ND \eqref{eqn:normalized_deviation} with $x_{k,T+h}^{\textrm{ref}} = \hat{x}_{k,T+h}$ as reference values, for $h \in H = \{2,\dots,\tau\}$.
This measures the relative discrepancy between the forecasts $(\hat{x}_{T+1}, \dots, \hat{x}_{T+\tau})$ and $(\hat{x}_{T+2}', \dots, \hat{x}_{T+\tau+1}')$, respectively based on $\bx$ and $(\bx;\tilde{x}_{T+1})$, at common indices.
The lower this value is, the more consistent the forecasts are, before and after the arrival of adversarial observation $\tilde{x}_{T+1}$.
We scale each input series $\bx$ before applying noises; i.e., $\bx$ is replaced by $\bx/S_{\bx}$ for some $S_{\bx}>0$ computed by the model to process each series within a consistent scale.

\paragraph{Results and discussion.}
Figure \ref{fig:translation_invariance_main} compares the metrics from vanilla DeepAR model, its smoothed version using Algorithm \ref{alg:future_smoothing} (labeled FS), the random-trained (RT) model as in the previous section, and its smoothed version (RT$+$FS), on each dataset.
The horizontal axes represent the adversarial parameter in a logarithmic scale $\log_{10}(1+\rho)$, and the vertical axes represent the relative ND.
The vanilla models already have a desirable behavior around $\rho=0$, but as $|\rho|$ grows, their forecast consistency is progressively broken.
On the other hand, RT models tend to be more resilient compared to vanilla models. 
FS provides statistically significant improvement in forecast consistency to vanilla and RT models for large values of $\rho$ in many cases (see Table~\ref{tab:translation_invariance_appendix} in the appendix).
We elicit a message similar to that of Section~\ref{subsection:experiments_adversarial_perturbation}; RT$+$FS is theoretically well-supported, and often achieves superior empirical performance as well.
We provide Table~\ref{tab:translation_invariance_appendix} containing all experiment results in the appendix.

\section{CONCLUSION}
\label{section:conclusion}

In this paper, we study robustness in the context of probabilistic time series forecasting.
We provide a framework of robust forecasting based on randomized smoothing with theoretical certificates, 
and display empirical effectiveness of the randomizing framework against two distinct types of input perturbations.

The formal treatment of robustness for probabilistic forecasting models is still only at its beginning.
We anticipate that the topic of robust probabilistic forecasting allows for multiple interesting and promising directions of future work, including establishing tighter theoretical guarantees or more extensive empirical study with broader classes of transformations.

\newpage

\section*{Acknowledgements}
TY and EKR were supported by the National Research Foundation of Korea (NRF) Grant funded by the Korean Government (MSIP) [No. 2017R1A5A1015626] and by the Samsung Science and Technology Foundation (Project Number SSTF-BA2101-02).
The authors would like to thank Nghia Hoang, Hilaf Hasson, Danielle Robinson, and Anoop Deoras in Amazon research for their fruitful feedback.

\bibliographystyle{plainnat}
\bibliography{ref}

\clearpage
\appendix

\thispagestyle{empty}

\onecolumn \makesupplementtitle

\section{PROOF OF THEORETICAL RESULTS}

\subsection{Randomized Smoothing of Deterministic Bounded Functions}
We first discuss some preliminary results on randomized smoothing of deterministic functions, which will be useful for the proofs in the subsequent sections.
Roughly speaking, smoothing \emph{any} bounded function results in a function with smoothness (Lipschitz continuity) property.
In the classification setup, \cite{cohen2019certified} provided a tight result bounding the certified radius, which is a lower bound on the norm of adversarial perturbation needed to incur incorrect classification.
Here we state its generalized version, proved in \cite{salman2019provably, levine2019certifiably}.

\begin{lemma}[\cite{salman2019provably, levine2019certifiably}]
\label{lemma:smoothed_function_Lipschitzness}
Given a function $h: \reals^d \to [0,1]$, define $$\hat{h}(\bx) := (h \asterisk \cN(0, I_d))(\bx) = \underset{\boldsymbol{Z}\sim \cN(0, I_d)}{\bbE}[h(\bx+\boldsymbol{Z})].$$
Then $\|\nabla \hat{h}(\bx)\| \le \phi(\Phi^{-1}(\hat{h}(\bx)))$, which implies that the mapping $\bx \mapsto \Phi^{-1}(\hat{h}(\bx))$ is 1-Lipschitz continuous.
\end{lemma}

To repeat the nomenclature, $\phi, \Phi$ are respectively the pdf and cdf of the standard normal distribution.
Note that if we denote $p_\sigma(\bz) = (2\pi\sigma^2)^{-d/2} \exp\left(-\|\bz\|^2/2\sigma^2\right)$ the pdf for the multivariate Gaussian distribution $\cN(0, \sigma^2 I_d)$ and $p(\bz) = p_1(\bz)$, then using the change of variables $\bz' = \bx + \bz$, we have
\begin{align}
    \nabla \hat{h}(\bx) &= \nabla_{\bx} \left( \int_{\bz \in \reals^d} h(\bx+\bz) p(\bz) \, d\bz \right) \nonumber \\
    &= \nabla_{\bx} \left( \int_{\bz' \in \reals^d} h(\bz') p(\bz'-\bx) \, d\bz' \right) \nonumber \\
    &= \int_{\bz' \in \reals^d} h(\bz') (\bz'-\bx) p(\bz'-\bx) \, d\bz' \nonumber \\
    &= \int_{\bz \in \reals^d} h(\bx+\bz) \bz p(\bz) \, d\bz . \label{eqn:smoothed_function_gradient}
\end{align}
The proof from \cite{salman2019provably} uses the argument that if $\bz \mapsto \psi(\bz)$ is a function such that $0 \le \psi \le 1$ and $$\int_{\bz\in\reals^d} \psi(\bz) p(\bz) \, d\bz = s,$$ then for any unit vector $\bu$, the following inequality holds:
\begin{align}
\label{eqn:supremum_over_fixed_expectation}
    \int_{\bz\in\reals^d} \psi(\bz) (\bu \cdot \bz) p(\bz) \, d\bz \le \int_{\bz\in\reals^d} \mathbf{1}_{\{\bz\in \reals^d \,|\, \bu \cdot \bz \ge -\Phi^{-1}(s)\}} (\bz) (\bu \cdot \bz) p(\bz) \, d\bz = \phi (\Phi^{-1}(s)).
\end{align}
Indeed, given the constrained budget on $\bbE_{\boldsymbol{Z}\sim \cN(0,I_d)} [\psi(\boldsymbol{Z})]$, one will maximize $\bbE_{\boldsymbol{Z}\sim \cN(0,I_d)} [(\bu \cdot \boldsymbol{Z})\psi(\boldsymbol{Z})]$ only by concentrating the mass $\psi(\bz)$ in the region with larger values of $\bu \cdot \bz$, i.e., on the set of the form $\{\bz\in\reals^d \,|\, \bu \cdot \bz \ge c\}$ for some $c \in \reals$.
Assuming $\bu = (1, 0, \dots, 0) \in \reals^d$ without loss of generality, we get
\begin{align*}
    \int_{\bz\in\reals^d} \mathbf{1}_{\{\bz\in \reals^d \,|\, \bu \cdot \bz \ge c\}}(\bz) p(\bz) \, d\bz = \int_c^\infty \phi(z_1) \, dz_1 = s \,\iff\, c = -\Phi^{-1}(s).
\end{align*}

\subsection{Proof of Theorem \ref{thm:smoothing}}
Note that in \eqref{eqn:thm_bound}, we abused the notation and written the $W_1$ distance in terms of cumulative distribution functions, as one can directly quantify the $W_1$ distance in terms of cdf's: if $F, G$ are respectively the cdf of a real-valued random variable and $\mu, \nu$ are the corresponding distributions, then
\begin{align}
\label{eqn:W1_dist_cdf}
    W_1 (\mu, \nu) = \int_0^1 |F^{-1} (p) - G^{-1}(p)| \, dp  = \int_{-\infty}^\infty |F(r) - G(r)| \, dr.
\end{align}
In the case of smoothed scalar predictors $g_{\sigma}$, respectively evaluated at $\bx$ and $\bx+\bdelta$, we have
\begin{align*}
    W_1(G_{\bx,\sigma}, G_{\bx+\bdelta, \sigma}) &= \int_{-\infty}^\infty \left| G_{\bx,\sigma}(r) - G_{\bx+\bdelta,\sigma}(r)\right| \, dr \\
    &= \int_{-\infty}^\infty \left| \int_{\bz \in \reals^T} F_{\bx + \bz}(r) p_\sigma(\bz) \, d\bz - \int_{\bz \in \reals^T} F_{\bx + \bdelta + \bz}(r) p_\sigma(\bz) \, d\bz \right| \, dr \\
    &= \int_{-\infty}^\infty \left| \int_{\bz\in\reals^T} F_{\bz}(r) (p_\sigma(\bz-\bx) - p_\sigma(\bz-\bx-\bdelta)) \, d\bz \right| \, dr.
\end{align*}
Now note that 
\begin{align*}
    \int_{\bz\in\reals^T} F_{\bz}(r) (p_\sigma(\bz-\bx) - p_\sigma(\bz-\bx-\bdelta)) \, d\bz & = \int_{\bz\in\reals^T} F_{\bz}(r) \int_{0}^1 \nabla p_\sigma (\bz - \bx - t\bdelta) \cdot (-\bdelta) \, dt \, d\bz \\
    & = \int_0^1 \int_{\bz\in \reals^T} F_{\bz}(r) \, \left( \bdelta \cdot \frac{\bz - \bx - t\bdelta}{\sigma^2} \right) \, p_\sigma (\bz - \bx - t\bdelta) \, d\bz \, dt \\
    & = \frac{1}{\sigma} \int_0^1 \int_{\bz' \in \reals^T} F_{\bx + t\bdelta + \sigma \bz'} (r) \, (\bdelta \cdot \bz') \, p(\bz') \, d\bz' \, dt
\end{align*}
where in the last line we make the change of variables $\bz' = \frac{\bz - \bx - t \bdelta}{\sigma}$.
Because the mapping $\bz' \mapsto F_{\bx + t\bdelta + \sigma\bz'}(r)$ is a function $\reals^T \to [0,1]$, by \eqref{eqn:supremum_over_fixed_expectation} we have
\begin{align}
    \label{eqn:bound_line_integral}
    \left| \int_{\bz\in\reals^T} F_{\bz}(r) (p_\sigma(\bz-\bx) - p_\sigma(\bz-\bx-\bdelta)) \, d\bz \right| & \le \frac{1}{\sigma} \int_0^1 \left| \int_{\bz' \in \reals^T} F_{\bx + t\bdelta + \sigma \bz'} (r) \, (\bdelta \cdot \bz') \, p(\bz') \, d\bz' \right| \, dt \\
    & \le \frac{1}{\sigma} \int_0^1 \|\bdelta\| \cdot \left\| \int_{\bz' \in \reals^T} F_{\bx + t\bdelta + \sigma\bz'} (r) \bz' p(\bz') \, d\bz' \right\| \, dt \nonumber \\
    & \le \frac{1}{\sigma} \int_0^1 \|\bdelta\| \cdot \phi \left(\Phi^{-1}\left( \int_{\bz'\in \reals^T} F_{\bx + t\bdelta + \sigma \bz'}(r) \, p(\bz') \, d\bz' \right) \right) \, dt \nonumber \\
    & = \frac{1}{\sigma} \int_0^1 \|\bdelta\| \cdot \phi(\Phi^{-1}(G_{\bx + t\bdelta, \sigma} (r) )) \, dt. \nonumber
\end{align}
Therefore, provided that $\int_{-\infty}^\infty \phi(\Phi^{-1}(G_{\bx', \sigma} (r) )) \, dr < C$ for $\bx'$ near $\bx$ for some $C > 0$, one can apply the dominated convergence theorem to obtain
\begin{align*}
    \limsup_{\|\bdelta\|\to 0} \frac{W_1(G_{\bx,\sigma}, G_{\bx+\bdelta, \sigma})}{\|\bdelta\|} & \le \limsup_{\|\bdelta\|\to 0} \int_{-\infty}^\infty \frac{1}{\sigma} \int_0^1 \phi(\Phi^{-1}(G_{\bx+t\bdelta,\sigma}(r))) \,dt \, dr \\
    & = \int_{-\infty}^\infty \frac{1}{\sigma} \int_0^1 \lim_{\|\bdelta\|\to 0} \phi(\Phi^{-1}(G_{\bx+t\bdelta,\sigma}(r))) \,dt \, dr \\
    & = \frac{1}{\sigma} \int_{-\infty}^{\infty} \phi\left(\Phi^{-1}(G_{\bx,\sigma}(r))\right) \, dr \\
    & = \mathrm{Ro}(\bx; \sigma),
\end{align*}
where we applied the fact that $\lim_{\|\bdelta\|\to 0} G_{\bx+t\bdelta, \sigma}(r) = G_{\bx}(r)$ by Lemma \ref{lemma:smoothed_function_Lipschitzness}.

\subsection{Proof of Lemma~\ref{lemma:finiteness}}

For $r < 0$, we proceed similarly as we bounded \eqref{eqn:bound_line_integral}, but bound the integrand in a different way:
\begin{align*}
    & \left| \int_{\bz\in\reals^T} \prob[f_j(\bz) \le r] \, (p_\sigma(\bz-\bx) - p_\sigma(\bz-\bx-\bdelta)) \, d\bz \right| \\
    & \quad \quad \le \frac{1}{\sigma} \int_0^1 \left| \int_{\bz \in \reals^T} \prob[f_j(\bx + t\delta + \sigma\bz) \le r] \, (\bdelta \cdot \bz) \, p(\bz) \, d\bz \right| \, dt \\
    & \quad \quad \le \frac{1}{\sigma} \int_0^1 \|\bdelta\| \, \int_{\bz \in \reals^T} \prob[f_j(\bx + t\delta + \sigma\bz) \le r] \, \|\bz\| \, p(\bz) \, d\bz  \, dt
\end{align*}
for $j=1,\dots,\tau$.
Note that
\begin{align*}
    \int_{\bz\in\reals^T} p_\sigma (\bz-\bx) - p_\sigma (\bz-\bx-\bdelta) \, d\bz = 0,
\end{align*}
so for $r > 0$, we can write
\begin{align}
    & \left| \int_{\bz\in\reals^T} \prob[f_j(\bz) \le r] \, (p_\sigma(\bz-\bx) - p_\sigma(\bz-\bx-\bdelta)) \, d\bz \right| \nonumber \\
    & \quad \quad = \left| \int_{\bz\in\reals^T} (\prob[f_j(\bz) \le r] - 1) (p_\sigma(\bz-\bx) - p_\sigma(\bz-\bx-\bdelta)) \, d\bz \right| \nonumber \\
    & \quad \quad = \left| \int_{\bz\in\reals^T} \prob[f_j(\bz) > r] \, (p_\sigma(\bz-\bx) - p_\sigma(\bz-\bx-\bdelta)) \, d\bz \right| \nonumber \\
    & \quad \quad \le \frac{1}{\sigma} \int_0^1 \|\bdelta\| \, \int_{\bz \in \reals^T} \prob[f_j(\bx + t\delta + \sigma\bz) > r] \, \|\bz\| \, p(\bz) \, d\bz  \, dt. \nonumber
\end{align}
Therefore, if we denote $F_{j,\bx}(r) = \prob[f_j(\bx) \le r]$ and $G_{j,\bx,\sigma}(r) = \int_{\bz\in\reals^T} F_{j,\bx+\bz}(r) p_{\sigma}(\bz) \, d\bz$,
\begin{align*}
    W_1(G_{j,\bx,\sigma}, G_{j,\bx+\bdelta, \sigma}) &= \int_{-\infty}^\infty \left| \int_{\bz\in\reals^T} F_{j,\bz}(r) (p_\sigma(\bz-\bx) - p_\sigma(\bz-\bx-\bdelta)) \, d\bz \right| \, dr \\
    &= \int_{-\infty}^0 \left| \int_{\bz\in\reals^T} F_{j,\bz}(r) (p_\sigma(\bz-\bx) - p_\sigma(\bz-\bx-\bdelta)) \, d\bz \right| \, dr \\
    & \quad \quad + \int_0^\infty \left| \int_{\bz\in\reals^T} F_{j,\bz}(r) (p_\sigma(\bz-\bx) - p_\sigma(\bz-\bx-\bdelta)) \, d\bz \right| \, dr \\
    & = \frac{\|\bdelta\|}{\sigma} \int_0^\infty \int_0^1 \int_{\bz \in \reals^T} \left( \prob[f_j(\bx + t\bdelta + \sigma\bz) \le -r] + \prob[f_j(\bx + t\bdelta + \sigma\bz) > r] \right) \|\bz\| \, p(\bz) \, d\bz \, dt \, dr \\
    & \le \frac{\|\bdelta\|}{\sigma} \left( \int_0^\infty \varphi(r) \, dr \right) \left( \int_{\bz\in \reals^T} \|\bz\| p(\bz)\, d\bz \right).
\end{align*}
This shows that the smoothed forecaster $g(\bx) = (g_1(\bx), \dots, g_\tau(\bx))$ is $C\eta-\eta$ robust in the given sense, where
\begin{align*}
    C = \frac{1}{\sigma} \left( \int_0^\infty \varphi(r) \, dr \right) \left( \int_{\bz\in \reals^T} \|\bz\| p(\bz)\, d\bz \right) < \infty.
\end{align*}

\paragraph{Remark.}
The constant $C$ may be chosen more tightly; for example, applying the bound in the above proof for $r>R$ with some large $R$, and using the original bound of Theorem~\ref{thm:smoothing} for $r\le R$ would result in
\begin{align*}
    C = \frac{1}{\sigma} \left( \int_{-R}^R \phi(\Phi^{-1}(G_{\bx,\sigma}(r)))\, dr + \int_{R}^\infty \varphi(r) \, dr \int_{\bz\in \reals^T} \|\bz\| p(\bz)\, d\bz \right) ,
\end{align*}
which can be smaller.

\subsection{Proof of Lemma~\ref{lemma:distributions}}

Suppose that there is $M_1, M_2 > 0$ such that $\left|\bbE[f(\bx)]\right| \le M_1$ and $\mathrm{Var}[f(\bx)] \le M_2$ holds for all $\bx \in \reals^T$.
Then for any $\bx\in\reals^T$ and $r > 0$, 
\begin{align*}
    r^2 \mathrm{Pr}[|f(\bx)| \ge r] \le \bbE [f(\bx)^2] = \mathrm{Var}[f(\bx)] + \left|\bbE[f(\bx)]\right|^2 \le M_1^2 + M_2 .
\end{align*}
Thus one can simply take $\varphi(r) = 1$ for $r \in [0,1]$ and $\varphi(r) = \frac{M_1^2 + M_2}{r^2}$ for $r>1$.

\subsection{Proof of Corollary~\ref{cor:future_smoothing}}

Denote $g_{\sigma}(\bx) = (Y_1, Y_2, \dots)$ and $g_{\sigma}(\bx; \tilde{\bx}_{T+1:T+k}) = (Y_{k+1}', Y_{k+2}', \dots)$.
Denote the probability measures corresponding to the random variables $Y_h, Y_h'$ respectively by $\mu_h, \mu_h'$.
When $\bx \in \reals^T$ is fixed, we can view $f^{(h)}(\bx; y_1,\dots,y_{h-1})$ as a function of $(y_1,\dots,y_{h-1}) \in \reals^h$, for each $h=1,2,\dots$.
Because we have
\begin{align*}
    \prob\left[ \left|f^{(h)}(\bx; y_1, \dots, y_{h-1})\right| \ge r\right] \le \varphi (r),
\end{align*}
we can apply Corollary~\ref{cor:finiteness} with $f^{(h)}\left(\bx; \cdot \right) : \reals^{h-1} \to \reals$, which implies that there exists $C_h > 0$ such that
\begin{align*}
    g_\sigma^{(h)} (\bx; y_1, \dots, y_{h-1}) = \underset{\boldsymbol{\zeta} \sim \cN(0, \sigma^2 I_{h-1})}{\bbE} \left[ f(\bx_{1:T}; y_1 + \zeta_1, \cdots, y_{h-1} + \zeta_{h-1}) \right]
\end{align*}
satisfies 
\begin{align*}
    W_1 \left( G_{\by',\sigma}^{(h)}, G_{\by,\sigma}^{(h)} \right) \le C_h \left\| \by' - \by \right\|_2
\end{align*}
for any $\by = (y_1,\dots,y_{h-1})$ and $\by' = (y_1',\dots,y_{h-1}')$, where $G_{\by,\sigma}^{(k)}$ is the cdf for the random variable $g_\sigma^{(k)}(\by)$ and $W_1$ distance between the cdfs denotes the $W_1$ distance between the corresponding probability measures by abuse of notation.
Applying the above bound in the case $h=k+1$, $\by = \hat{\bx}_{T+1:T+k}$ and $\by' = \tilde{\bx}_{T+1:T+k}$ gives
\begin{align*}
    W_1(\mu_{k+1}', \mu_{k+1}) = W_1 \left( G_{\tilde{\bx}_{T+1:T+k},\sigma}^{(k)}, G_{\hat{\bx}_{T+1:T+k},\sigma}^{(k)} \right) \le C_k \left\| \tilde{\bx}_{T+1:T+k} - \hat{\bx}_{T+1:T+k} \right\|_2 
\end{align*}
because $Y_{k+1} = g_{\sigma}^{(k)}(\bx;\hat{\bx}_{T+1:T+k})$ and $Y_{k+1}' = g_{\sigma}^{(k)}(\bx;\tilde{\bx}_{T+1:T+k})$.
In particular, this bounds the difference between the mean point forecasts:
\begin{align*}
    \left| \overline{Y}'_{k+1} - \overline{Y}_{k+1} \right| = \left| \hat{x}'_{T+k+1} - \hat{x}_{T+k+1} \right| & = \left| \bbE \left[ g_\sigma^{(k)}(\bx;\tilde{\bx}_{T+1:T+k})] - \bbE[g_\sigma^{(k)}(\bx;\hat{\bx}_{T+1:T+k}) \right] \right| \\
    & \le \underset{\chi:1\text{-Lipschitz}}{\sup} \bbE\left[ \chi\left( g_\sigma^{(k)}(\bx;\tilde{\bx}_{T+1:T+k}) \right) \right] -  \bbE\left[\chi\left(g_\sigma^{(k)}(\bx;\hat{\bx}_{T+1:T+k})\right)\right] \\
    & = W_1 \left( G_{\by',\sigma}^{(h)}, G_{\by,\sigma}^{(h)} \right) \\
    & \le C_k \left\| \tilde{\bx}_{T+1:T+k} - \hat{\bx}_{T+1:T+k} \right\|_2 .
\end{align*}
Then we obtain
\begin{align*}
    \left\| \left( \tilde{\bx}_{T+1:T+k}; \hat{x}'_{T+k+1} \right) - \left( \hat{\bx}_{T+1:T+k}; \hat{x}_{T+k+1} \right) \right\|_2 
    \le (1 + C_k) \left\| \tilde{\bx}_{T+1:T+k} - \hat{\bx}_{T+1:T+k} \right\|_2 ,
\end{align*}
which again implies
\begin{align*}
    W_1(\mu_{k+2}', \mu_{k+2}) & = W_1 \left( G_{(\tilde{\bx}_{T+1:T+k};\hat{\bx}_{T+k+1}'),\sigma}^{(k)}, G_{(\hat{\bx}_{T+1:T+k};\hat{\bx}_{T+k+1}),\sigma}^{(k)} \right) \\
    & \le C_{k+1} \left\| \left( \tilde{\bx}_{T+1:T+k}; \hat{x}'_{T+k+1} \right) - \left( \hat{\bx}_{T+1:T+k}; \hat{x}_{T+k+1} \right) \right\|_2 \\
    & \le C_{k+1} (1+C_k) \left\| \tilde{\bx}_{T+1:T+k} - \hat{\bx}_{T+1:T+k} \right\|_2 .
\end{align*}
Repeating the same argument, we see that
\begin{align*}
    T_{\cY}(g_{\sigma}(\bx)) = T_{\cY}(Y_1, Y_2, \dots ) = (Y_{k+1}, Y_{k+2}, \dots) \approx (Y_{k+1}', Y_{k+2}', \dots) = g_{\sigma} \left( \bx; \tilde{\bx}_{T+1:T+k} \right) = g_{\sigma} \left( T_{\cX}(\bx) \right)
\end{align*}
in the sense that
\begin{align*}
    W_1(\mu'_{k+j}, \mu_{k+j}) = O\left( \left\| \tilde{\bx}_{T+1:T+k} - \hat{\bx}_{T+1:T+k} \right\|_2 \right) = O\left( d_{\cX} (\bx;T_{\cX}) \right)
\end{align*}
holds for each $j=1,2,\dots$, which completes the proof.

\section{EXPERIMENTAL DETAILS}

We use the time series forecasting library GluonTS \citep{alexandrov2020gluonts} to configure and train the vanilla and random-trained DeepAR models, and to perform randomized smoothing using them.
For the experiments in Sections~\ref{subsection:experiments_adversarial_perturbation} and \ref{subsection:experiments_translation_invariance}, we use the standard prediction lengths $\tau=24$ for the Electricity and Traffic datasets, $\tau=30$ for the Exchange Rate dataset, and $\tau=14$ for the M4-Daily dataset.
The context lengths are set to $4\tau$ for all cases, and all the other model hyperparameters are set to default values within the GluonTS implementation.
The training of all baseline models (with or without data augmentation with random noises) uses batch size $128$ and is run for $50$ epochs.
We use 100 sample paths from each baseline and smoothed model to perform adversarial attack and generate forecasts.
The code for the experiments is available at \url{https://github.com/tetrzim/robust-probabilistic-forecasting}.

\section{ADDITIONAL EXPERIMENT RESULTS}

\begin{figure}[ht]
\captionsetup[subfigure]{justification=centering}
    \centering
    \begin{subfigure}{0.33\linewidth}
    \centering
    \includegraphics[scale=0.32]{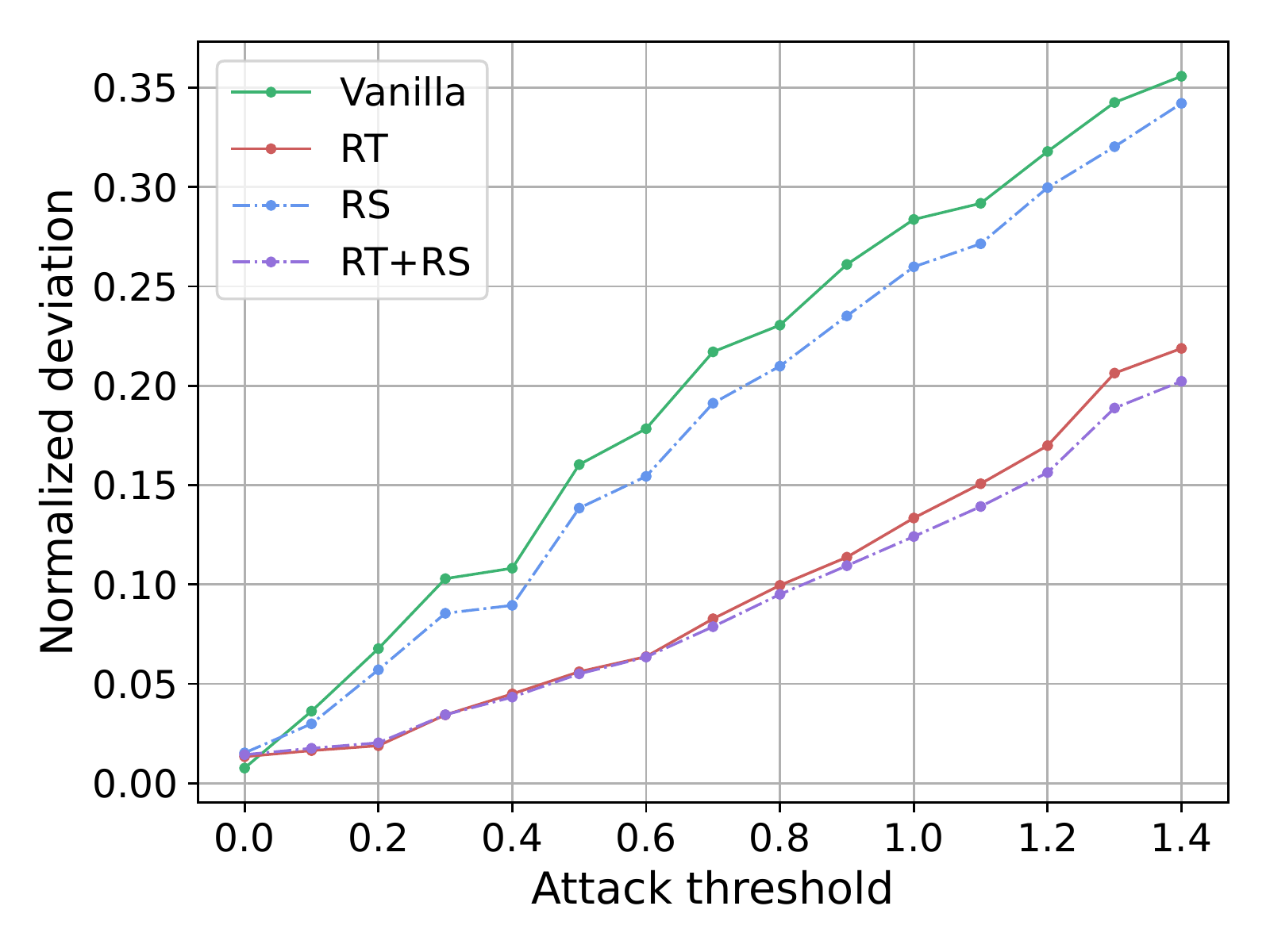}
    \caption{Exchange Rate, $\tau = 30$, $\{H\} = \{1\}$}
    \label{subfig:exchange_Rate_tau_30_H_1}
    \end{subfigure}
    \hspace{-.2cm}
    \begin{subfigure}{0.33\linewidth}
    \centering
    \includegraphics[scale=0.32]{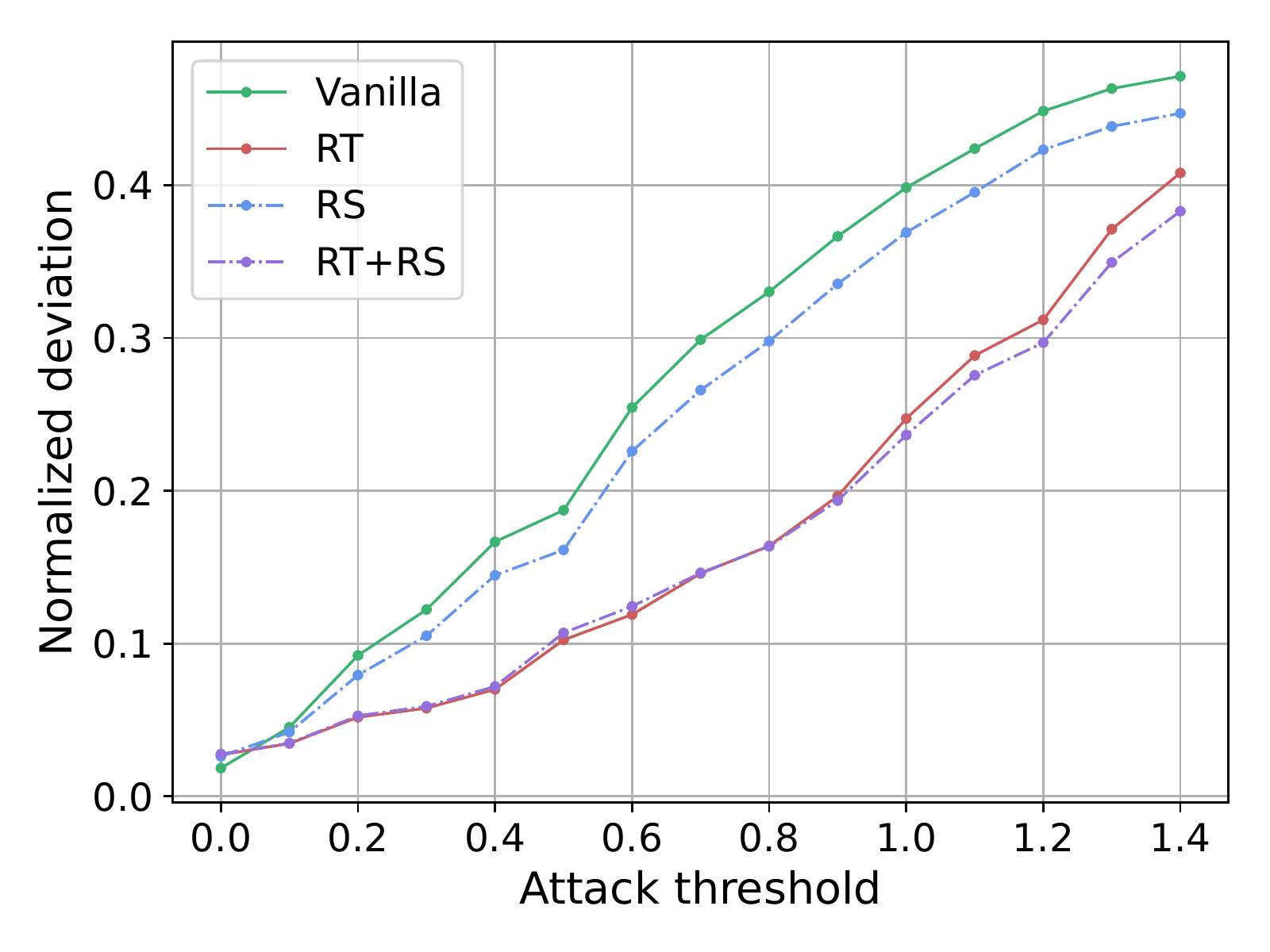}
    \caption{M4-Daily, $\tau = 14$, $\{H\} = \{1\}$}
    \label{subfig:m4_daily_tau_14_H_1}
    \end{subfigure}
    \hspace{-.2cm}
    \begin{subfigure}{0.33\linewidth}
    \centering
    \includegraphics[scale=0.32]{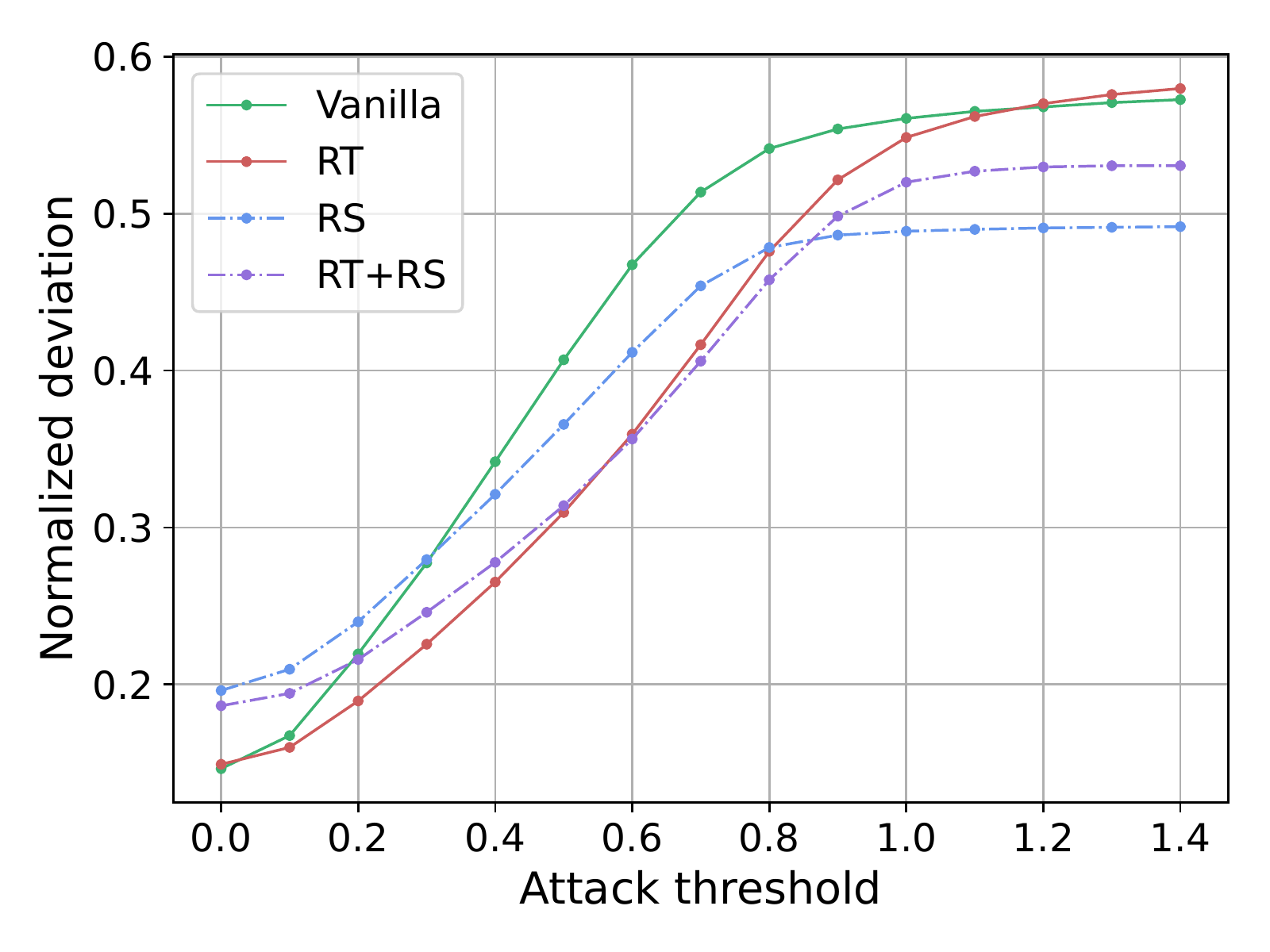}
    \caption{Traffic, $\tau = 24$, $\{H\} = \{1\}$}
    \label{subfig:traffic_tau_30_H_1}
    \end{subfigure}\\
    \vspace{.3cm}
    \begin{subfigure}{0.33\linewidth}
    \centering
    \includegraphics[scale=0.32]{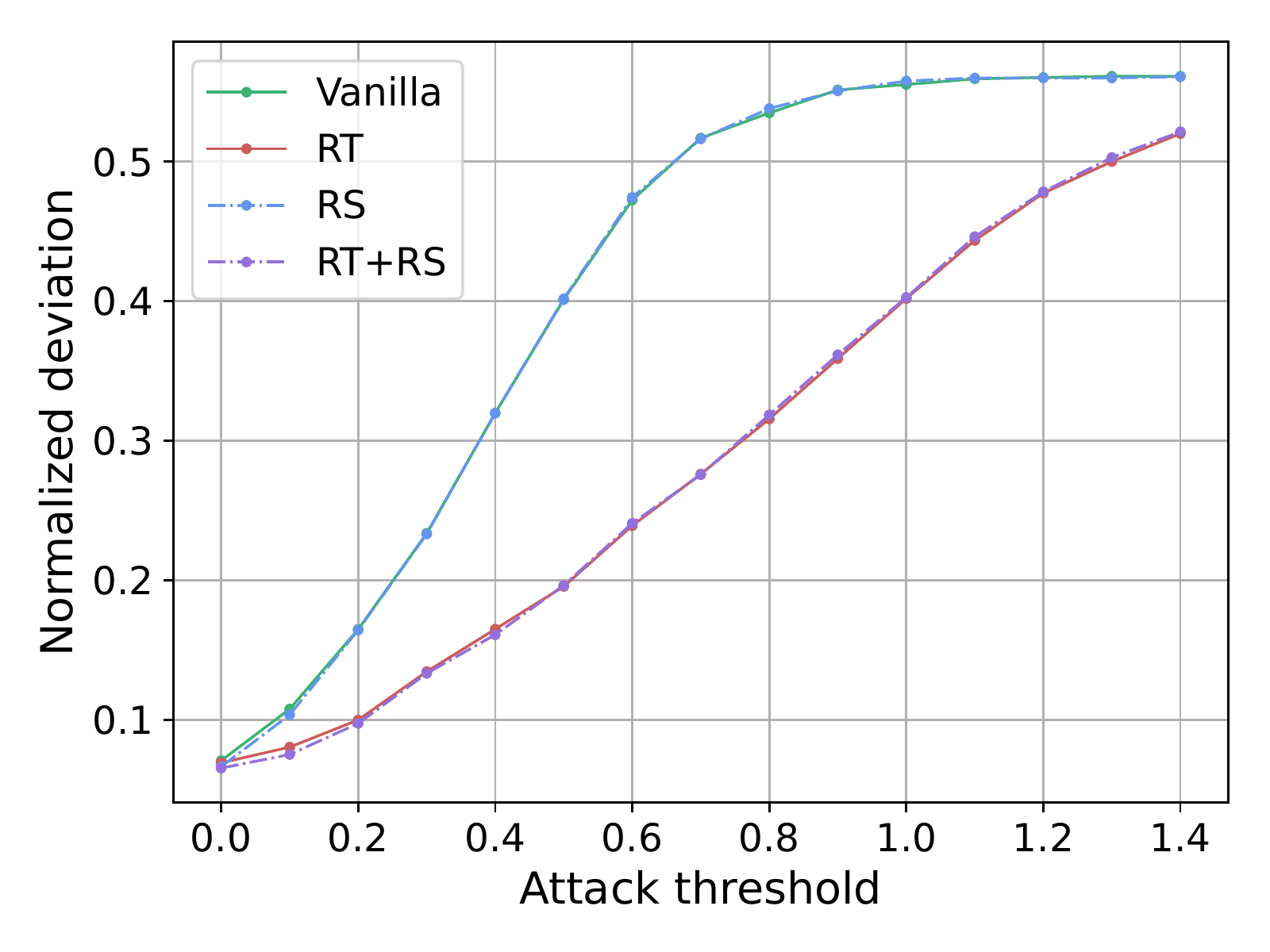}
    \caption{Electricity, $\tau = 24$, $\{H\} = \{1\}$}
    \label{subfig:electricity_tau_24_H_1}
    \end{subfigure}
    \begin{subfigure}{0.33\linewidth}
    \centering
    \includegraphics[scale=0.32]{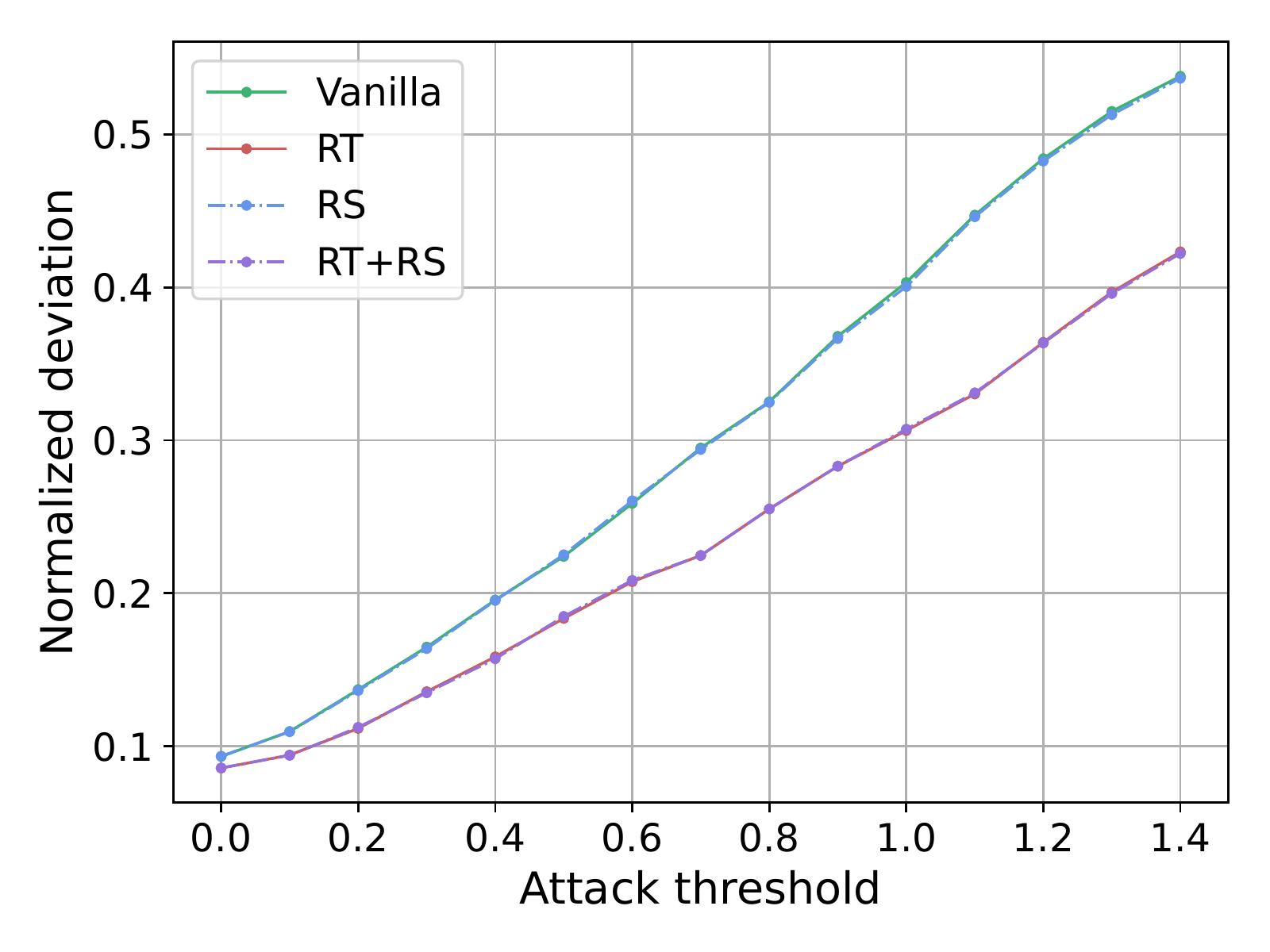}
    \caption{Electricity, $\tau = 24$, $\{H\} = \{\tau\}$}
    \label{subfig:electricity_tau_24_H_24}
    \end{subfigure}
    \caption{$\mathrm{ND}_H$ from DeepAR models on different datasets under adversarial attacks with respect to relative $l_2$ norm. Smoothing of baseline models uses $\sigma = 0.1$ for the Electricity dataset and $\sigma = 0.5$ for other datasets. Randomized training uses $\sigma_{\mathrm{tr}} = 0.1$.}
    \label{fig:adversarial_accuracy_appendix}
\end{figure}

\begin{table}[ht]
\centering
\caption{Mean and standard deviation of ND on clean test set for \emph{all} prediction indices over 10 runs.}
    \begin{tabular}{cccc}
    \toprule
                  & Vanilla  & \makecell{Random-trained \\($\sigma_{\mathrm{tr}} = 0.1$)} \\ \midrule
    Exchange Rate & 0.024$\pm$0.008 & \textbf{0.018}$\pm$0.001 \\ \midrule 
    Traffic       & 0.131$\pm$0.006 & \textbf{0.127}$\pm$0.003 \\ \midrule
    Electricity   & 0.075$\pm$0.010 & \textbf{0.067}$\pm$0.005 \\
    \bottomrule
    \end{tabular}
\label{tab:randomizing_and_clean_accuracy}
\end{table}

Figure~\ref{fig:adversarial_accuracy_appendix}, together with Figure~\ref{fig:adversarial_accuracy_main} and Table~\ref{tab:adversarial_perturbation_appendix}, shows the prediction accuracy (measured in terms of ND on the attacked indices) of DeepAR models under the adversarial attack of \cite{dang2020adversarial} with respect to relative $l_2$ norms.
Table~\ref{tab:translation_invariance_appendix}, together with Figure~\ref{fig:translation_invariance_main}, shows the relative ND on overlapping indices under supplement of a noisy observation.

Table~\ref{tab:randomizing_and_clean_accuracy} compares ND on the test set without adversarial attack, measured over all prediction indices (that is, $\mathrm{ND}_H$ \eqref{eqn:normalized_deviation} with $H=\{1,\dots,\tau\}$).
Each row indicates that random-trained (RT) models have attained better accuracy compared to the corresponding vanilla model.
Because the training of DeepAR involves windowing over multiple time intervals, and we apply noise to every observation available in the training data, randomized training noises both input and output (label) values.
It is well-known that randomization of training inputs is positively related to generalization \citep{sietsma1991creating, matsuoka1992noise, holmstrom1992using, bishop1995training}, 
as observed by \cite{zhang2007neural} in the context of time series forecasting.
On the other hand, some recent results \citep{blanc2020implicit, damian2021label} provided theoretical study on the implicit bias of optimization algorithms with label noise towards well-generalizing minima.
However, we are not aware of prior works that particularly studied the effect of training with label noising on forecasting accuracy.

\begin{table}[ht]
    \caption{Mean and standard deviation of $\mathrm{ND}_H$ from DeepAR models on different datasets under adversarial attacks, measured over 10 independent runs. The \textbf{*} symbols for RS models indicate statistically significant improvement against the corresponding baselines according to the Wilcoxon signed-rank test.}
    \label{tab:adversarial_perturbation_appendix}
    \vspace{.5cm}
    \begin{subtable}[ht]{\textwidth}
    \centering
    \caption{Exchange Rate}
    \label{tab:exchange_rate}
    \begin{tabular}{c|c|cccc}
    \toprule
    $H$ & $\eta$ & Vanilla & \makecell{RS \\ ($\sigma = 0.5$)} & \makecell{RT \\ ($\sigma_{\textrm{tr}} = 0.1$)} & \makecell{RT + RS \\ ($\sigma_{\mathrm{tr}} = 0.1, \sigma = 0.5$)} \\ \midrule
    \multirow{8}{*}{$\{1\}$} & 0 & \textbf{0.008}$\,\pm\,$0.002 & 0.015$\,\pm\,$0.003 & 0.013$\,\pm\,$0.002 & 0.014$\,\pm\,$0.003 \\
    & 0.2 & 0.068$\,\pm\,$0.013 & 0.057$\,\pm\,$0.010\textsuperscript{\textbf{*}} & \textbf{0.019}$\,\pm\,$0.002 & \textbf{0.020}$\,\pm\,$0.002 \\
    & 0.4 & 0.108$\,\pm\,$0.021 & 0.090$\,\pm\,$0.014\textsuperscript{\textbf{*}} & 0.045$\,\pm\,$0.005 & \textbf{0.043}$\,\pm\,$0.006 \\
    & 0.6 & 0.178$\,\pm\,$0.026 & 0.154$\,\pm\,$0.019\textsuperscript{\textbf{*}} & \textbf{0.064}$\,\pm\,$0.012 & \textbf{0.063}$\,\pm\,$0.008 \\
    & 0.8 & 0.231$\,\pm\,$0.031 & 0.210$\,\pm\,$0.031\textsuperscript{\textbf{*}} & 0.100$\,\pm\,$0.014 & \textbf{0.095}$\,\pm\,$0.011 \\
    & 1.0 & 0.284$\,\pm\,$0.038 & 0.260$\,\pm\,$0.038\textsuperscript{\textbf{*}} & 0.133$\,\pm\,$0.037 & \textbf{0.124}$\,\pm\,$0.028\textsuperscript{\textbf{*}} \\
    & 1.2 & 0.318$\,\pm\,$0.048 & 0.300$\,\pm\,$0.046\textsuperscript{\textbf{*}} & 0.170$\,\pm\,$0.044 & \textbf{0.156}$\,\pm\,$0.035\textsuperscript{\textbf{*}} \\
    & 1.4 & 0.356$\,\pm\,$0.052 & 0.342$\,\pm\,$0.054\textsuperscript{\textbf{*}} & 0.219$\,\pm\,$0.029 & \textbf{0.202}$\,\pm\,$0.024\textsuperscript{\textbf{*}} \\
    \midrule
    \multirow{8}{*}{$\{\tau\}$} &  0 & 0.036$\,\pm\,$0.014 & 0.043$\,\pm\,$0.017 & \textbf{0.022}$\,\pm\,$0.002 & \textbf{0.023}$\,\pm\,$0.005 \\
    & 0.2 & 0.065$\,\pm\,$0.016 & 0.067$\,\pm\,$0.017 & \textbf{0.023}$\,\pm\,$0.002 & \textbf{0.023}$\,\pm\,$0.007 \\
    & 0.4 & 0.111$\,\pm\,$0.035 & 0.099$\,\pm\,$0.029\textsuperscript{\textbf{*}} & \textbf{0.025}$\,\pm\,$0.002 & \textbf{0.026}$\,\pm\,$0.003 \\
    & 0.6 & 0.164$\,\pm\,$0.054 & 0.142$\,\pm\,$0.044\textsuperscript{\textbf{*}} & 0.035$\,\pm\,$0.021 & \textbf{0.031}$\,\pm\,$0.006 \\
    & 0.8 & 0.194$\,\pm\,$0.057 & 0.174$\,\pm\,$0.050\textsuperscript{\textbf{*}} & 0.054$\,\pm\,$0.033 & \textbf{0.049}$\,\pm\,$0.020 \\
    & 1.0 & 0.210$\,\pm\,$0.060 & 0.192$\,\pm\,$0.057\textsuperscript{\textbf{*}} & 0.117$\,\pm\,$0.058 & \textbf{0.099}$\,\pm\,$0.042\textsuperscript{\textbf{*}} \\
    & 1.2 & 0.230$\,\pm\,$0.059 & 0.216$\,\pm\,$0.057\textsuperscript{\textbf{*}} & 0.173$\,\pm\,$0.066 & \textbf{0.149}$\,\pm\,$0.052\textsuperscript{\textbf{*}} \\
    & 1.4 & 0.243$\,\pm\,$0.062 & 0.230$\,\pm\,$0.060\textsuperscript{\textbf{*}} & 0.215$\,\pm\,$0.038 & \textbf{0.191}$\,\pm\,$0.034\textsuperscript{\textbf{*}} \\
    \bottomrule
    \end{tabular}
    \end{subtable}\\
    \vspace{1cm}\\
    \begin{subtable}[ht]{\textwidth}
    \centering
    \caption{M4-Daily}
    \label{tab:m4_daily_tau_14}
    \begin{tabular}{c|c|cccc}
    \toprule
    $H$ & $\eta$ & Vanilla & \makecell{RS \\ ($\sigma = 0.5$)} & \makecell{RT \\ ($\sigma_{\textrm{tr}} = 0.1$)} & \makecell{RT + RS \\ ($\sigma_{\mathrm{tr}} = 0.1, \sigma = 0.5$)} \\ \midrule
    \multirow{8}{*}{$\{1\}$}  & 0 & \textbf{0.018}$\,\pm\,$0.002 & 0.026$\,\pm\,$0.001 & 0.027$\,\pm\,$0.001 & 0.028$\,\pm\,$0.002 \\
    & 0.2 & 0.092$\,\pm\,$0.004 & 0.079$\,\pm\,$0.003\textsuperscript{\textbf{*}} & \textbf{0.052}$\,\pm\,$0.007 & \textbf{0.053}$\,\pm\,$0.007 \\
    & 0.4 & 0.167$\,\pm\,$0.021 & 0.145$\,\pm\,$0.018\textsuperscript{\textbf{*}} & \textbf{0.070}$\,\pm\,$0.012 & 0.072$\,\pm\,$0.011 \\
    & 0.6 & 0.254$\,\pm\,$0.024 & 0.226$\,\pm\,$0.021\textsuperscript{\textbf{*}} & \textbf{0.119}$\,\pm\,$0.011 & 0.124$\,\pm\,$0.011 \\
    & 0.8 & 0.330$\,\pm\,$0.034 & 0.298$\,\pm\,$0.032\textsuperscript{\textbf{*}} & \textbf{0.164}$\,\pm\,$0.028 & \textbf{0.164}$\,\pm\,$0.025 \\
    & 1.0 & 0.398$\,\pm\,$0.041 & 0.369$\,\pm\,$0.038\textsuperscript{\textbf{*}} & 0.247$\,\pm\,$0.037 & \textbf{0.236}$\,\pm\,$0.030\textsuperscript{\textbf{*}} \\
    & 1.2 & 0.449$\,\pm\,$0.027 & 0.423$\,\pm\,$0.027\textsuperscript{\textbf{*}} & 0.312$\,\pm\,$0.041 & \textbf{0.297}$\,\pm\,$0.034\textsuperscript{\textbf{*}} \\
    & 1.4 & 0.471$\,\pm\,$0.017 & 0.447$\,\pm\,$0.015\textsuperscript{\textbf{*}} & 0.408$\,\pm\,$0.034 & \textbf{0.383}$\,\pm\,$0.033\textsuperscript{\textbf{*}} \\
    \midrule
    \multirow{8}{*}{$\{\tau\}$} & 0 & 0.062$\,\pm\,$0.011 & 0.065$\,\pm\,$0.009 & \textbf{0.056}$\,\pm\,$0.003 & \textbf{0.055}$\,\pm\,$0.005 \\
    & 0.2 & 0.114$\,\pm\,$0.019 & 0.103$\,\pm\,$0.016\textsuperscript{\textbf{*}} & \textbf{0.062}$\,\pm\,$0.003 & \textbf{0.061}$\,\pm\,$0.006 \\
    & 0.4 & 0.162$\,\pm\,$0.020 & 0.141$\,\pm\,$0.017\textsuperscript{\textbf{*}} & \textbf{0.074}$\,\pm\,$0.006 & \textbf{0.075}$\,\pm\,$0.010 \\
    & 0.6 & 0.247$\,\pm\,$0.027 & 0.209$\,\pm\,$0.026\textsuperscript{\textbf{*}} & \textbf{0.084}$\,\pm\,$0.007 & \textbf{0.085}$\,\pm\,$0.011 \\
    & 0.8 & 0.321$\,\pm\,$0.044 & 0.275$\,\pm\,$0.041\textsuperscript{\textbf{*}} & 0.135$\,\pm\,$0.058 & \textbf{0.130}$\,\pm\,$0.052 \\
    & 1.0 & 0.376$\,\pm\,$0.052 & 0.327$\,\pm\,$0.049\textsuperscript{\textbf{*}} & 0.219$\,\pm\,$0.096 & \textbf{0.194}$\,\pm\,$0.075\textsuperscript{\textbf{*}} \\
    & 1.2 & 0.415$\,\pm\,$0.056 & 0.366$\,\pm\,$0.053\textsuperscript{\textbf{*}} & 0.394$\,\pm\,$0.085 & \textbf{0.333}$\,\pm\,$0.072\textsuperscript{\textbf{*}} \\
    & 1.4 & 0.442$\,\pm\,$0.045 & \textbf{0.392}$\,\pm\,$0.042\textsuperscript{\textbf{*}} & 0.493$\,\pm\,$0.025 & 0.415$\,\pm\,$0.040\textsuperscript{\textbf{*}} \\
    \bottomrule
    \end{tabular}
    \end{subtable}
\end{table}

\begin{table}[ht]
    \addtocounter{table}{-1}
    \caption{\textit{(Continued)} Mean and standard deviation of $\mathrm{ND}_H$ from DeepAR models on different datasets under adversarial attacks, measured over 10 independent runs. The \textbf{*} symbols for RS models indicate statistically significant improvement against the corresponding baselines according to the Wilcoxon signed-rank test.}
    \label{tab:adversarial_perturbation_appendix_2}
    \centering
    \vspace{.5cm}
    \begin{subtable}[ht]{\textwidth}
    \addtocounter{subtable}{2}
    \centering
    \caption{Traffic}
    \label{tab:traffic_tau_24}
    \begin{tabular}{c|c|cccc}
    \toprule
    $H$ & $\eta$ & Vanilla & \makecell{RS \\ ($\sigma = 0.5$)} & \makecell{RT \\ ($\sigma_{\textrm{tr}} = 0.1$)} & \makecell{RT + RS \\ ($\sigma_{\mathrm{tr}} = 0.1, \sigma = 0.5$)} \\ \midrule
    \multirow{8}{*}{$\{1\}$}  & 0 & \textbf{0.146}$\,\pm\,$0.002 & 0.196$\,\pm\,$0.004 & 0.149$\,\pm\,$0.003 & 0.186$\,\pm\,$0.007 \\
    & 0.2 & 0.220$\,\pm\,$0.003 & 0.240$\,\pm\,$0.005 & \textbf{0.190}$\,\pm\,$0.003 & 0.216$\,\pm\,$0.006 \\
    & 0.4 & 0.342$\,\pm\,$0.005 & 0.321$\,\pm\,$0.004\textsuperscript{\textbf{*}} & \textbf{0.265}$\,\pm\,$0.005 & 0.278$\,\pm\,$0.006 \\
    & 0.6 & 0.467$\,\pm\,$0.007 & 0.412$\,\pm\,$0.005\textsuperscript{\textbf{*}} & \textbf{0.359}$\,\pm\,$0.007 & \textbf{0.356}$\,\pm\,$0.013 \\
    & 0.8 & 0.541$\,\pm\,$0.003 & 0.478$\,\pm\,$0.004\textsuperscript{\textbf{*}} & 0.476$\,\pm\,$0.014 & \textbf{0.458}$\,\pm\,$0.024\textsuperscript{\textbf{*}} \\
    & 1.0 & 0.561$\,\pm\,$0.002 & \textbf{0.489}$\,\pm\,$0.006\textsuperscript{\textbf{*}} & 0.548$\,\pm\,$0.006 & 0.520$\,\pm\,$0.013\textsuperscript{\textbf{*}} \\
    & 1.2 & 0.568$\,\pm\,$0.002 & \textbf{0.491}$\,\pm\,$0.007\textsuperscript{\textbf{*}} & 0.570$\,\pm\,$0.003 & 0.530$\,\pm\,$0.011\textsuperscript{\textbf{*}} \\
    & 1.4 & 0.573$\,\pm\,$0.003 & \textbf{0.492}$\,\pm\,$0.007\textsuperscript{\textbf{*}} & 0.580$\,\pm\,$0.003 & 0.531$\,\pm\,$0.011\textsuperscript{\textbf{*}} \\
    \midrule
    \multirow{8}{*}{$\{\tau\}$} & 0 & 0.183$\,\pm\,$0.007 & 0.229$\,\pm\,$0.012 & \textbf{0.179}$\,\pm\,$0.005 & 0.213$\,\pm\,$0.014 \\
    & 0.2 & 0.212$\,\pm\,$0.008 & 0.241$\,\pm\,$0.011 & \textbf{0.200}$\,\pm\,$0.005 & 0.223$\,\pm\,$0.014 \\
    & 0.4 & 0.261$\,\pm\,$0.009 & 0.267$\,\pm\,$0.011 & \textbf{0.240}$\,\pm\,$0.006 & 0.245$\,\pm\,$0.014 \\
    & 0.6 & 0.309$\,\pm\,$0.011 & 0.296$\,\pm\,$0.011 & 0.279$\,\pm\,$0.007 & \textbf{0.272$\,\pm\,$0.015} \\
    & 0.8 & 0.367$\,\pm\,$0.011 & 0.340$\,\pm\,$0.013\textsuperscript{\textbf{*}} & 0.324$\,\pm\,$0.008 & \textbf{0.307}$\,\pm\,$0.014\textsuperscript{\textbf{*}} \\
    & 1.0 & 0.435$\,\pm\,$0.014 & 0.389$\,\pm\,$0.014\textsuperscript{\textbf{*}} & 0.378$\,\pm\,$0.011 & \textbf{0.348}$\,\pm\,$0.015\textsuperscript{\textbf{*}} \\
    & 1.2 & 0.507$\,\pm\,$0.015 & 0.450$\,\pm\,$0.015\textsuperscript{\textbf{*}} & 0.444$\,\pm\,$0.013 & \textbf{0.409}$\,\pm\,$0.016\textsuperscript{\textbf{*}} \\
    & 1.4 & 0.561$\,\pm\,$0.014 & 0.491$\,\pm\,$0.015\textsuperscript{\textbf{*}} & 0.505$\,\pm\,$0.015 & \textbf{0.464}$\,\pm\,$0.017\textsuperscript{\textbf{*}} \\
    \bottomrule
    \end{tabular}
    \end{subtable}\\
    \vspace{1cm}
    \begin{subtable}[ht]{\textwidth}
    \centering
    \caption{Electricity}
    \label{tab:electricity_tau_24}
    \begin{tabular}{c|c|cccc}
    \toprule
    $H$ & $\eta$ & Vanilla & \makecell{RS \\ ($\sigma = 0.1$)} & \makecell{RT \\ ($\sigma_{\textrm{tr}} = 0.1$)} & \makecell{RT + RS \\ ($\sigma_{\mathrm{tr}} = 0.1, \sigma = 0.1$)} \\ \midrule
    \multirow{8}{*}{$\{1\}$}  & 0 & 0.071$\,\pm\,$0.003 & \textbf{0.066}$\,\pm\,$0.003\textsuperscript{\textbf{*}} & 0.069$\,\pm\,$0.009 & \textbf{0.065}$\,\pm\,$0.008\textsuperscript{\textbf{*}} \\
    & 0.2 & 0.165$\,\pm\,$0.005 & 0.164$\,\pm\,$0.004 & 0.100$\,\pm\,$0.009 & \textbf{0.097}$\,\pm\,$0.008 \\
    & 0.4 & 0.320$\,\pm\,$0.012 & 0.320$\,\pm\,$0.009 & 0.165$\,\pm\,$0.011 & \textbf{0.161}$\,\pm\,$0.009\textsuperscript{\textbf{*}} \\
    & 0.6 & 0.472$\,\pm\,$0.013 & 0.474$\,\pm\,$0.011 & \textbf{0.239}$\,\pm\,$0.011 & \textbf{0.241}$\,\pm\,$0.011 \\
    & 0.8 & 0.535$\,\pm\,$0.009 & 0.538$\,\pm\,$0.011 & \textbf{0.316}$\,\pm\,$0.019 & \textbf{0.318}$\,\pm\,$0.017 \\
    & 1.0 & 0.555$\,\pm\,$0.013 & 0.558$\,\pm\,$0.009 & \textbf{0.402}$\,\pm\,$0.019 & \textbf{0.403}$\,\pm\,$0.019 \\
    & 1.2 & 0.560$\,\pm\,$0.011 & 0.559$\,\pm\,$0.011 & \textbf{0.477}$\,\pm\,$0.017 & \textbf{0.478}$\,\pm\,$0.017 \\
    & 1.4 & 0.559$\,\pm\,$0.014 & 0.561$\,\pm\,$0.011 & \textbf{0.520}$\,\pm\,$0.015 & \textbf{0.521}$\,\pm\,$0.015 \\
    \midrule
    \multirow{8}{*}{$\{\tau\}$} & 0 & 0.093$\,\pm\,$0.018 & 0.093$\,\pm\,$0.016 & \textbf{0.086}$\,\pm\,$0.013 & \textbf{0.086}$\,\pm\,$0.014 \\
    & 0.2 & 0.137$\,\pm\,$0.021 & 0.136$\,\pm\,$0.021 & \textbf{0.112}$\,\pm\,$0.014 & \textbf{0.112}$\,\pm\,$0.014 \\
    & 0.4 & 0.196$\,\pm\,$0.023 & 0.195$\,\pm\,$0.024 & \textbf{0.158}$\,\pm\,$0.015 & \textbf{0.157}$\,\pm\,$0.015 \\
    & 0.6 & 0.259$\,\pm\,$0.029 & 0.260$\,\pm\,$0.028 & \textbf{0.208}$\,\pm\,$0.018 & \textbf{0.208}$\,\pm\,$0.017 \\
    & 0.8 & 0.325$\,\pm\,$0.035 & 0.325$\,\pm\,$0.033 & \textbf{0.255}$\,\pm\,$0.017 & \textbf{0.255}$\,\pm\,$0.014 \\
    & 1.0 & 0.403$\,\pm\,$0.043 & 0.401$\,\pm\,$0.041 & \textbf{0.306}$\,\pm\,$0.019 & \textbf{0.307}$\,\pm\,$0.018 \\
    & 1.2 & 0.484$\,\pm\,$0.049 & 0.483$\,\pm\,$0.048 & \textbf{0.364}$\,\pm\,$0.018 & \textbf{0.364}$\,\pm\,$0.019 \\
    & 1.4 & 0.538$\,\pm\,$0.046 & 0.537$\,\pm\,$0.048 & \textbf{0.423}$\,\pm\,$0.019 & \textbf{0.422}$\,\pm\,$0.021 \\
    \bottomrule
    \end{tabular}
    \end{subtable}
\end{table}

\newpage

\begin{table}[ht]
    \caption{Mean and standard deviation of relative $\mathrm{ND}$ from DeepAR models on different datasets under supplement of noisy observation with adversarial parameter $\rho$, measured over 10 independent runs.\\ The \textbf{*} symbols for FS models indicate statistically significant improvement against the corresponding baselines according to the Wilcoxon signed-rank test.}
    \label{tab:translation_invariance_appendix}
    \centering
    \vspace{.5cm}
    \begin{subtable}[ht]{\textwidth}
    \centering
    \caption{Exchange Rate}
    \label{tab:translation_exchange_rate}
    \begin{tabular}{c|cccc}
    \toprule
    $\rho$ & Vanilla & \makecell{FS \\ ($\sigma = 1.0$)} & \makecell{RT \\ ($\sigma_{\textrm{tr}} = 0.1$)} & \makecell{RT + FS \\ ($\sigma_{\textrm{tr}} = 0.1$, $\sigma = 1.0$)} \\ \midrule
    -0.9 & 0.160$\,\pm\,$0.003 & 0.093$\,\pm\,$0.005\textsuperscript{\textbf{*}} & \textbf{0.034}$\,\pm\,$0.003 & \textbf{0.033}$\,\pm\,$0.005 \\
    -0.5 & 0.091$\,\pm\,$0.001 & 0.057$\,\pm\,$0.002\textsuperscript{\textbf{*}} & \textbf{0.019}$\,\pm\,$0.001 & 0.021$\,\pm\,$0.002 \\
    0    & \textbf{0.003}$\,\pm\,$0.000 & 0.024$\,\pm\,$0.000 & 0.012$\,\pm\,$0.000 & 0.014$\,\pm\,$0.000 \\
    0.5  & 0.082$\,\pm\,$0.002 & 0.057$\,\pm\,$0.001\textsuperscript{\textbf{*}} & \textbf{0.019}$\,\pm\,$0.002 & \textbf{0.020}$\,\pm\,$0.001 \\
    1.0  & 0.150$\,\pm\,$0.007 & 0.101$\,\pm\,$0.005\textsuperscript{\textbf{*}} & 0.036$\,\pm\,$0.007 & \textbf{0.034}$\,\pm\,$0.005\textsuperscript{\textbf{*}} \\
    2.0  & 0.238$\,\pm\,$0.019 & 0.178$\,\pm\,$0.012\textsuperscript{\textbf{*}} & 0.073$\,\pm\,$0.019 & \textbf{0.063}$\,\pm\,$0.012\textsuperscript{\textbf{*}} \\
    4.0  & 0.315$\,\pm\,$0.030 & 0.263$\,\pm\,$0.019\textsuperscript{\textbf{*}} & 0.125$\,\pm\,$0.030 & \textbf{0.109}$\,\pm\,$0.019\textsuperscript{\textbf{*}} \\
    9.0  & 0.391$\,\pm\,$0.034 & 0.347$\,\pm\,$0.024\textsuperscript{\textbf{*}} & 0.190$\,\pm\,$0.034 & \textbf{0.178}$\,\pm\,$0.024\textsuperscript{\textbf{*}} \\
    \bottomrule
    \end{tabular}
    \end{subtable}\\
    \vspace{1cm}
    \begin{subtable}[ht]{\textwidth}
    \centering
    \caption{M4-Daily}
    \label{tab:translation_m4_daily}
    \begin{tabular}{c|cccc}
    \toprule
    $\rho$ & Vanilla & \makecell{FS \\ ($\sigma = 1.0$)} & \makecell{RT \\ ($\sigma_{\textrm{tr}} = 0.1$)} & \makecell{RT + FS \\ ($\sigma_{\textrm{tr}} = 0.1$, $\sigma = 1.0$)} \\ \midrule
    -0.9 & 0.272$\,\pm\,$0.012 & 0.173$\,\pm\,$0.014\textsuperscript{\textbf{*}} & \textbf{0.095}$\,\pm\,$0.012 & 0.098$\,\pm\,$0.014 \\
    -0.5 & 0.157$\,\pm\,$0.002 & 0.099$\,\pm\,$0.007\textsuperscript{\textbf{*}} & \textbf{0.047}$\,\pm\,$0.002 & 0.052$\,\pm\,$0.007 \\
    0    & \textbf{0.007}$\,\pm\,$0.000 & 0.037$\,\pm\,$0.002 & 0.013$\,\pm\,$0.000 & 0.021$\,\pm\,$0.002 \\
    0.5  & 0.151$\,\pm\,$0.002 & 0.101$\,\pm\,$0.007\textsuperscript{\textbf{*}} & \textbf{0.046}$\,\pm\,$0.002 & 0.052$\,\pm\,$0.007 \\
    1.0  & 0.269$\,\pm\,$0.008 & 0.185$\,\pm\,$0.015\textsuperscript{\textbf{*}} & \textbf{0.097}$\,\pm\,$0.008 & 0.099$\,\pm\,$0.015 \\
    2.0  & 0.416$\,\pm\,$0.035 & 0.319$\,\pm\,$0.034\textsuperscript{\textbf{*}} & 0.215$\,\pm\,$0.035 & \textbf{0.196}$\,\pm\,$0.034\textsuperscript{\textbf{*}} \\
    4.0  & 0.558$\,\pm\,$0.073 & 0.483$\,\pm\,$0.066\textsuperscript{\textbf{*}} & 0.405$\,\pm\,$0.073 & \textbf{0.363}$\,\pm\,$0.066\textsuperscript{\textbf{*}} \\
    9.0  & 0.723$\,\pm\,$0.090 & 0.681$\,\pm\,$0.093\textsuperscript{\textbf{*}} & 0.619$\,\pm\,$0.090 & \textbf{0.597}$\,\pm\,$0.093\textsuperscript{\textbf{*}} \\
    \bottomrule
    \end{tabular}
    \end{subtable}
\end{table}

\newpage

\begin{table}[ht]
    \addtocounter{table}{-1}
    \caption{\textit{(Continued)} Mean and standard deviation of relative $\mathrm{ND}$ from DeepAR models on different datasets under supplement of noisy observation with adversarial parameter $\rho$, measured over 10 independent runs. The \textbf{*} symbols for FS models indicate statistically significant improvement against the corresponding baselines according to the Wilcoxon signed-rank test.}
    \label{tab:translation_invariance_appendix_2}
    \centering
    \vspace{.5cm}
    \begin{subtable}[ht]{\textwidth}
    \addtocounter{subtable}{2}
    \centering
    \caption{Traffic}
    \label{tab:translation_traffic}
    \begin{tabular}{c|cccc}
    \toprule
    $\rho$ & Vanilla & \makecell{FS \\ ($\sigma = 0.1$)} & \makecell{RT \\ ($\sigma_{\textrm{tr}} = 0.1$)} & \makecell{RT + FS \\ ($\sigma_{\textrm{tr}} = 0.1$, $\sigma = 0.1$)} \\ \midrule
    -0.9 & 0.141$\,\pm\,$0.008 & 0.139$\,\pm\,$0.008 & 0.102$\,\pm\,$0.008 & \textbf{0.100}$\,\pm\,$0.008\textsuperscript{\textbf{*}} \\
    -0.5 & 0.084$\,\pm\,$0.003 & 0.085$\,\pm\,$0.003 & \textbf{0.058}$\,\pm\,$0.003 & \textbf{0.058}$\,\pm\,$0.003 \\
    0 & 0.037$\,\pm\,$0.001 & 0.039$\,\pm\,$0.001 & \textbf{0.034}$\,\pm\,$0.001 & \textbf{0.033}$\,\pm\,$0.001\textsuperscript{\textbf{*}} \\
    0.5 & 0.068$\,\pm\,$0.002 & 0.068$\,\pm\,$0.002\textsuperscript{\textbf{*}} & 0.048$\,\pm\,$0.002 & \textbf{0.046}$\,\pm\,$0.002\textsuperscript{\textbf{*}} \\
    1.0 & 0.096$\,\pm\,$0.002 & 0.094$\,\pm\,$0.003\textsuperscript{\textbf{*}} & 0.065$\,\pm\,$0.002 & \textbf{0.063}$\,\pm\,$0.003\textsuperscript{\textbf{*}} \\
    2.0 & 0.121$\,\pm\,$0.004 & 0.119$\,\pm\,$0.004\textsuperscript{\textbf{*}} & 0.089$\,\pm\,$0.004 & \textbf{0.086}$\,\pm\,$0.004\textsuperscript{\textbf{*}} \\
    4.0 & 0.139$\,\pm\,$0.007 & 0.137$\,\pm\,$0.007\textsuperscript{\textbf{*}} & 0.109$\,\pm\,$0.007 & \textbf{0.106}$\,\pm\,$0.007\textsuperscript{\textbf{*}} \\
    9.0 & 0.170$\,\pm\,$0.010 & 0.167$\,\pm\,$0.009\textsuperscript{\textbf{*}} & 0.140$\,\pm\,$0.010 & \textbf{0.137}$\,\pm\,$0.009\textsuperscript{\textbf{*}} \\
    \bottomrule
    \end{tabular}
    \end{subtable}\\
    \vspace{1cm}
    \begin{subtable}[ht]{\textwidth}
    \centering
    \caption{Electricity}
    \label{tab:translation_electricity}
    \begin{tabular}{c|cccc}
    \toprule
    $\rho$ & Vanilla & \makecell{FS \\ ($\sigma = 0.5$)} & \makecell{RT \\ ($\sigma_{\textrm{tr}} = 0.1$)} & \makecell{RT + FS \\ ($\sigma_{\textrm{tr}} = 0.1$, $\sigma = 0.5$)} \\ \midrule
    -0.9 & 0.065$\,\pm\,$0.001 & 0.078$\,\pm\,$0.002 & \textbf{0.034}$\,\pm\,$0.001 & 0.039$\,\pm\,$0.002 \\
    -0.5 & 0.042$\,\pm\,$0.001 & 0.061$\,\pm\,$0.002 & \textbf{0.025}$\,\pm\,$0.001 & 0.032$\,\pm\,$0.002 \\
    0 & \textbf{0.017}$\,\pm\,$0.001 & 0.044$\,\pm\,$0.002 & 0.019$\,\pm\,$0.001 & 0.025$\,\pm\,$0.002 \\
    0.5 & 0.047$\,\pm\,$0.001 & 0.058$\,\pm\,$0.002 & \textbf{0.025}$\,\pm\,$0.001 & 0.031$\,\pm\,$0.002 \\
    1.0 & 0.088$\,\pm\,$0.002 & 0.087$\,\pm\,$0.003 & \textbf{0.037}$\,\pm\,$0.002 & 0.041$\,\pm\,$0.003 \\
    2.0 & 0.170$\,\pm\,$0.004 & 0.147$\,\pm\,$0.005\textsuperscript{\textbf{*}} & \textbf{0.066}$\,\pm\,$0.004 & 0.069$\,\pm\,$0.005 \\
    4.0 & 0.295$\,\pm\,$0.018 & 0.246$\,\pm\,$0.016\textsuperscript{\textbf{*}} & 0.136$\,\pm\,$0.018 & \textbf{0.134}$\,\pm\,$0.016\textsuperscript{\textbf{*}} \\
    9.0 & 0.435$\,\pm\,$0.055 & 0.379$\,\pm\,$0.047\textsuperscript{\textbf{*}} & 0.291$\,\pm\,$0.055 & \textbf{0.276}$\,\pm\,$0.047\textsuperscript{\textbf{*}} \\
    \bottomrule
    \end{tabular}
    \end{subtable}
\end{table}

\end{document}

%% file: macros.tex
\DeclareMathOperator{\argmin}{\mathrm{argmin}}
\DeclareMathOperator{\argmax}{\mathrm{argmax}}

\newcommand{\prob}{\mathrm{Pr}\,}
\newcommand{\reals}{\mathbb{R}}

\newcommand{\bbE}{\mathbb{E}}

\newcommand{\bu}{\boldsymbol{u}}

\newcommand{\bx}{\boldsymbol{x}}
\newcommand{\by}{\boldsymbol{y}}
\newcommand{\bz}{\boldsymbol{z}}

\newcommand{\bY}{\mathbf{Y}}

\newcommand{\cF}{\mathcal{F}}
\newcommand{\cN}{\mathcal{N}}
\newcommand{\cP}{\mathcal{P}}
\newcommand{\cW}{\mathcal{W}}
\newcommand{\cX}{\mathcal{X}}
\newcommand{\cY}{\mathcal{Y}}
\newcommand{\cZ}{\mathcal{Z}}

\newcommand{\bdelta}{\boldsymbol{\delta}}